\documentclass{svjour3} 
\smartqed 
\usepackage[utf8]{inputenc}
\usepackage{amssymb}
\usepackage{listings}
\usepackage{mathrsfs}
\usepackage{times}
\usepackage{float}
\usepackage{color}
\usepackage{setspace}
\usepackage{color,soul}

\usepackage{amsmath,amsfonts,amsthm,eucal}
\usepackage{enumerate}
\usepackage[letterpaper,margin=3cm]{geometry}
\usepackage{float}
\usepackage[title]{appendix}
\usepackage{color}

\usepackage[
pdfstartview=XYZ,
bookmarks=true,
colorlinks=true,
linkcolor=blue,
urlcolor=blue,
citecolor=blue,
pdftex,
bookmarks=true,
linktocpage=true,   
hyperindex=true
]{hyperref}

\usepackage{lipsum}

\usepackage{lineno}

\usepackage[FIGBOTCAP,TABTOPCAP,bf,tight]{subfigure}

\usepackage{natbib}
\usepackage[pdftex]{graphicx}
\usepackage{epstopdf}
\usepackage{booktabs} 

\usepackage{tikz,mathpazo}
\usetikzlibrary{shapes.geometric, arrows}

\usepackage{caption}
\usepackage{mathtools}
\usepackage{etoolbox}

\newcommand{\tensor}[1]{\ensuremath{\boldsymbol{#1}}}

\newcommand{\eps}{\tensor{\epsilon}}
\newcommand{\sig}{\tensor{\sigma}}
\DeclareMathOperator{\grad}{\nabla^{\tensor x}}
\DeclareMathOperator{\diver}{\nabla^{\tensor x}\cdot}

\DeclareMathAlphabet{\mathpzc}{OT1}{pzc}{m}{it}

\DeclareMathOperator*{\argmin}{arg\,min}

\theoremstyle{remark}

\renewcommand{\vec}[1]{\ensuremath{\boldsymbol{#1}}}

\usepackage[super]{nth}
\usepackage{array}
\usepackage{multirow}
\newcommand{\fig}[1]{Fig. #1}
\newcommand{\Fig}[1]{Figure #1}

\newcommand{\mat}[1]{\ensuremath{\tensor{#1}}}

\newcolumntype{M}[1]{>{\centering\arraybackslash\hspace{0pt}}p{#1}}

\usepackage{algorithm}
\usepackage[noend]{algpseudocode}

\title{Manifold embedding data-driven mechanics}  

\begin{document}


\author{Bahador Bahmani      \and 
        WaiChing Sun 
}

\institute{Corresponding author: WaiChing Sun, PhD \at
Associate Professor, Department of Civil Engineering and Engineering Mechanics, 
 Columbia University , 
 614 SW Mudd, Mail Code: 4709, 
 New York, NY 10027
  Tel.: 212-854-3143, 
  Fax: 212-854-6267, 
  \email{wsun@columbia.edu}        
}

\date{Date: \today}

\maketitle

\begin{abstract}
This article introduces a new data-driven approach that leverages 
a manifold embedding generated by the invertible neural network to improve the robustness, efficiency, and accuracy of the constitutive-law-free simulations with limited data. 
We achieve this by training a deep neural network to globally map data from the constitutive manifold onto 
a lower-dimensional Euclidean vector space. 
As such, we establish the relation between the norm of the mapped Euclidean vector space and the metric
of the manifold and lead to a more physically consistent notion of distance for the material data. 
This treatment in return allows us to bypass the expensive combinatorial optimization, which may significantly speed up the model-free simulations when data are abundant and of high dimensions. 
Meanwhile, the learning of embedding also improves the robustness of the algorithm when the data 
is sparse or distributed unevenly in the parametric space. 
Numerical experiments are provided to demonstrate and measure the performance of the manifold embedding technique under different circumstances. Results obtained from the proposed method and those obtained via the classical energy norms are compared. 
\end{abstract}


\section{Introduction}
\label{intro}
Conventional computer simulations of mechanics problems often employ discretizations to convert boundary value problems into 
systems of equations and obtain the discretized solution from a solver. The boundary value problems 
are often split into two components, i.e., the constraints for the solution space-time domain that often derives from 
balance principles, and constitutive models that relate physical quantities derived from the solutions based on a combination 
of phenomenological observations and constraints derived from thermodynamics. 
Examples of these constitutive models are abundant in the literature of the last few centuries \citep{timoshenko1983history} even 
before the development of numerical methods for boundary value problems. 
Recently, the interest in deep learning has sparked a renewed interest in training neural networks to generate constitutive responses, 
(an idea that can track back to the last AI winter \citep{ghaboussi1991knowledge}) or replace classical solvers to generate numerical 
solutions of boundary value problems \citep{raissi2019physics}. 

Recently, \citet{kirchdoerfer2016data} introduce a new paradigm in which the constitutive models are bypassed. 
Instead, optimization problems that minimize the distance of material data and the manifold of the balance principles 
are introduced to generate solutions that satisfy both the balance principle and the phenomenological observations. 
As such, there is no need to compose equations for constitutive laws which could be an ambiguous, ad-hoc, and time-consuming process. 
The removal of the constitutive laws, therefore, bring in a new paradigm where the universal principles are enforced and the solutions 
are sought by finding the data points that are closest to the conservation manifold. 
However, one important question central to the success of this paradigm is the following-- how to properly measure the "distance" between the material database and conservation manifold?

Interestingly, there have been very few researches on the choice of the distance measure for the data-driven model-free approach.
In data-driven mechanics research, such as \citet{kirchdoerfer2016data} and \citet{kirchdoerfer2018data},  
the independent components of the input-output pairs (e.g., the strain-stress pair) of the constitutive laws are often used to constitute a product space, and the measure of distance relies on the equipped weighted energy norms. This idea is utilized not only for elasticity problems 
but also for multiphysics (e.g. \citet{bahmani2021kd} and higher-order continua (e.g. \citet{karapiperis2021data}).
Noticeably, \citet{leygue2018data} notes the implications of the choices of distance measure and report a dependence between
the energy-minimized solutions and the choice of the measure (see Fig. 6 of \citet{leygue2018data}). 
The underlying issue is not necessarily the choice of the distance measure but on the discrete nature 
of data point clouds. 
Without any prior assumption on the geometry, defining a unique length between two data points becomes impossible. For instance, the distance between two points on a flat surface and that on a curved surface are different. 
Assuming that the data are on a Euclidean space such as $\mathbb{R}^{12}$ for stress-strain data, 
could be feasible when the data distributed in the phase space is sufficiently dense or the data set itself is close to linear (and hence the difference between the length measured from the local tangential space and the constitutive manifold is minor). However, these conditions are not necessarily realistic.

\begin{figure}[h]
 \centering
{\includegraphics[width=0.9\textwidth]{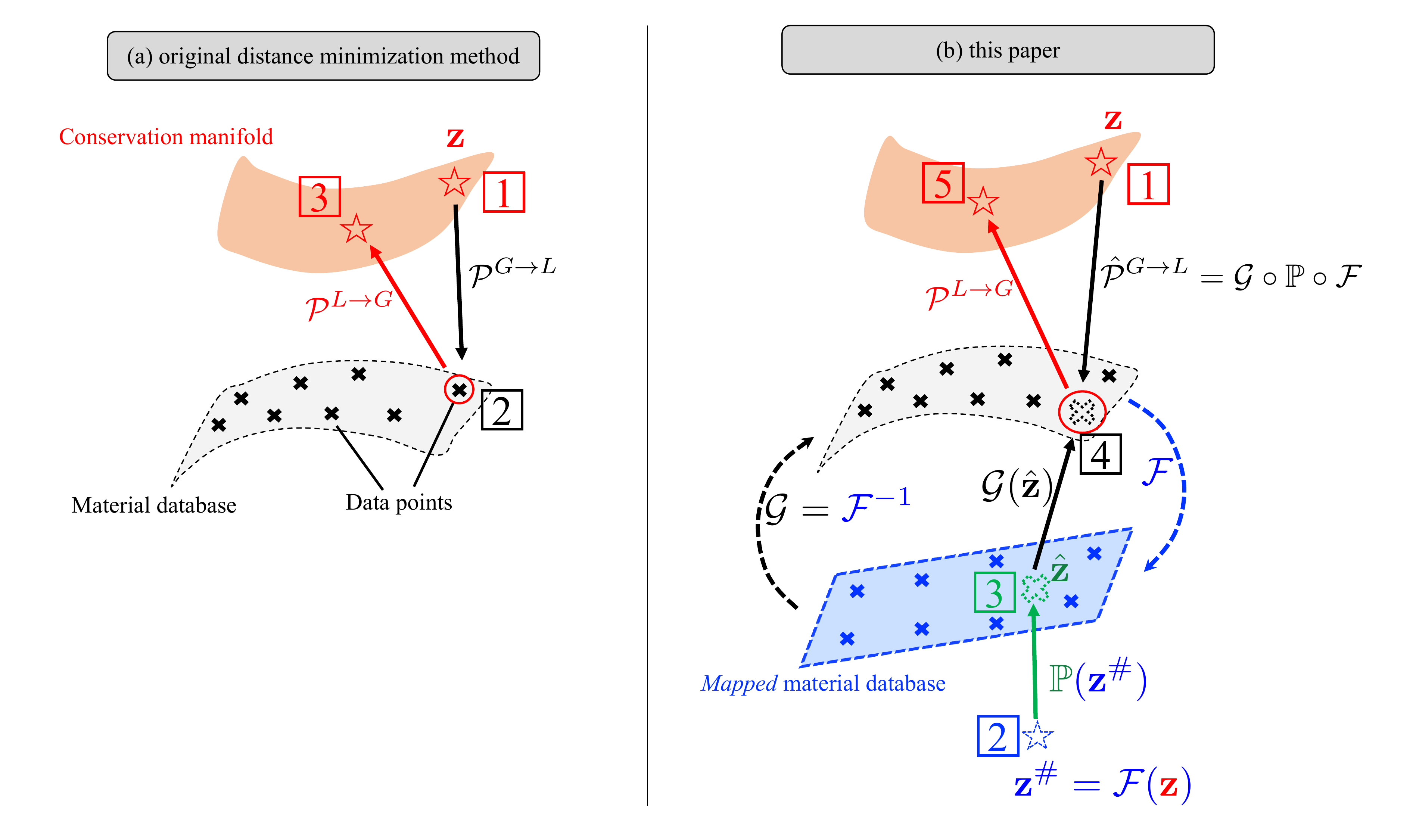}}
  \caption{Comparison between (a) the original distance minimization method and (b) the introduced method in this paper. The distance minimization algorithm iterates between two steps: first, local projection $\mathcal{P}^{G\mapsto L}$ from equilibrium state $\vec{z}$ numbered by 1 in (a) to material state numbered by 2; second, global projection $\mathcal{P}^{L\mapsto G}$ from material states to an updated equilibrium state numbered by 3. In our method, the local step $\hat{\mathcal{P}}^{G\mapsto L}$ in (b) is modified. We project an equilibrium state to the closest point on a previously constructed Euclidean space (shown by blue plane) corresponding to the material data space (shown by gray manifold).
   \label{fig::overview}}
\end{figure}

There have already been efforts that introduce resolutions to 
circumvent this spurious dependence. A common approach is to introduce \textit{locally linear embedding}. 
In \citet{ibanez2017data}, for instance, the data points used to measure the distance are factored by a weighting function such that each local linear patch around a data point can be mapped onto a lower-dimensional embedding space. 
In \citet{he2020physics}, locally convex material manifolds are also introduced
to achieve locally linear exactness with dimensional reductions. 
Meanwhile, \cite{kanno2021kernel} introduce a kernel-based method to extract the manifold of the constitutive responses \textit{globally}. By extracting the global constitutive manifold where all the data points belong to, a metric can be established via embedding. However, like the original 
data-driven approach where the computational cost may become significantly higher with an increasing number of data points, the complexity 
of the kernel method scales with the number of data points and hence can be expensive when handling a sufficiently large database even the dimensions of the data is relatively low (e.g., 1D constitutive laws).

In this paper, we introduce a neural network based \textbf{global} manifold embedding technique such that a proper distance measure corresponding to the manifold of the constitutive responses can be introduced 
for the data-driven mechanics simulations.
To achieve this goal, we consider the cases where we have collected data points from experiments
or direct numerical simulations. We then train an invertible neural network such that it may map all data points
on the constitutive manifold of path-independent materials onto a Euclidean space equipped with a Euclidean norm,
as shown in Figure \ref{fig::overview}. 

 This treatment is introduced to serve two purposes. 
First, we would like to improve the robustness of the model-free 
predictions when data are not sufficiently dense in the phase space. 
Second, we want to remove the computational bottleneck due to
the combinatorial optimization necessary for the closest 
point search in the data-driven paradigm. 
Our numerical experiments indicate that the global embedding technique may 
improve the robustness 
of the data-driven approach such that (1) it still functions reasonably well when data is sparse or not distributed with even density and (2) remains sufficiently efficient when data is abundant. 
Such an improvement is particularly important for high-consequence simulations 
where the robustness of the predictions becomes critical. 

The rest of the paper is organized as follows. We will first briefly review the model-free paradigm and outline the current state-of-the-art distance minimization techniques for the 
model-free approach. This review is followed by another brief outline on the manifold embedding and the 
manifold learning techniques commonly used for regression, classification, and computer vision problems. 
Other related works on predictions made on point cloud data in manifold and metric space are outlined in 
Section \ref{sec:otherworks}. 

The data-driven paradigm employs global and local minimization problems sequentially to locate the data points (on the constitutive manifold) closest to the conservation manifold. 
Our work employs a similar design, but the local minimization problem is reformulated in the mapped Euclidean space. 
To make the presentation self-contained,  Section \ref{sec::model-free} is provided
to outline the global optimization step for the model-free paradigm. We then revisit the local minimization problem in Section \ref{sec::new-local} and introduce the invertible neural network approach 
in Section \ref{sec::nn-embd} as the means to improve the efficiency of the local search (by replacing the combinatorial optimization with a projection performed in the mapped Euclidean space). Numerical examples 
are conducted to illustrate the ideas, verify the implementation and demonstrate and compare the performance of the manifold learning approach with the classical Euclidean counterpart in Section \ref{sec::numericalexample}. A few key observations are summarized in Section \ref{sec::conclusion}. 
Unless otherwise specified, we assume that all constitutive manifolds studied in this paper are Riemannian.

\subsection{Review on data-driven/model-free solid mechanics}
The model-free distance-minimization method (cf. \citet{kirchdoerfer2016data}) is a new paradigm that directly incorporates material databases as the replacements of explicit surrogate constitutive laws into the numerical simulations in a variationally consistent manner.  However, due to the minimal constraints 
imposed to generate numerical solutions, there must be sufficient data to ensure the distance between the 
data points and the conversation manifold are sufficiently close. Hence, the original formulation may 
exhibit sensitivity to noises and outliers  \citep{kirchdoerfer2017data,ibanez2017data,he2020physics,bahmani2021kd,kanno2021kernel}. In \citet{ibanez2017data}, the manifolds of the constitutive responses are constructed
in an unsupervised learning setting, and a locally linear embedding method proposed in 
\citet{roweis2000nonlinear} is used to recover the low dimensional characteristics of a material database locally.

Meanwhile, the robustness against outliers can be improved by a clustering scheme based on the maximum-entropy estimation \citep{kirchdoerfer2017data}. \citep{kanno2018data,kanno2018simple} formulate the distance minimization method based on a local kernel regression method (in an offline manner) to enhance the robustness in the noisy-data and limited-data scenarios. \citep{he2020physics} construct a locally linear subspace for points in a small neighborhood surrounding each material point (in an online manner). They perform convex interpolation on the constructed subspace to increase the robustness against noisy data.
However, locally linear embedding techniques are not suitable for the cases where sample density varies 
non-uniformly. Since the number of local sampling points that constitute the local linear data reconstructions may affect the approximated topology of the manifold (specifically in the limited data regime), it may require careful tuning.

\citet{kanno2021kernel} introduce a new formulation that constructs a global embedding parameterized by a kernel method in an offline manner. One drawback of kernel methods is the scalability issue with respect to the data size \citep{bishop2006pattern,zhang2012scaling}. Since kernel methods often operate on dense kernel matrices that scale quadratically with the data size, a database with 
many instances may therefore slow down the kernel method and make it less effective for big data applications. 
 \citet{eggersmann2021model} use a tensor-voting interpolation method locally inside a ball around a material point (in an offline manner) to increase the performance in the limited data regime and enhance the robustness against noise. \citep{he2021deep} introduces a nonlinear dimensionality reduction method based on the autoencoder architecture to perform noise-filtering.

The distance-minimization paradigm has also been extended for different applications such as elasto-dynamics \citep{kirchdoerfer2018data}, finite-strain elasticity \citep{nguyen2018data}, plasticity \citep{eggersmann2019model}, fracture mechanics \citep{carrara2020data}, geometrically exact beam theory \citep{gebhardt2020framework}, poroelasticity \citep{bahmani2021kd}, and micro-polar continuum \citep{karapiperis2021data}. Its scalability issue with respect to the data size is addressed in \citep{bahmani2021kd} where an isometric projection is introduced to efficiently organize the database via the kd-tree data structure that provides a logarithmic time complexity (in average) for nearest neighbor search. Other efficient data structures are also explored in \citep{eggersmann2021efficient}.

\subsection{Review on manifold embedding in machine learning}
The application of manifold learning in data-driven solid mechanics is shown previously, as discussed in the previous section. In this work, we aim to leverage the expressibility of neural networks to find a global Euclidean embedding with desirable properties. Hence, the first important question is the existence of such an embedding.

Whitney embedding theorem \citep{whitney1936differentiable,whitney1944self} proves the existence of a function $\mathcal{F}: \mathcal{M}\mapsto \mathbb{R}^n$ that maps a sufficiently smooth manifold $\mathcal{M}$ with intrinsic dimension $m$ to a vector (Euclidean) space of dimension $n\ge n_{\text{whitney}}$ where $n_{\text{whitney}} = 2m$, provided that $\mathcal{M}$ is not a real projective space. Nash proves the existence of such a mapping  even with the isometry restriction for $n \ge n_{\text{nash}}$ where $n_{\text{nash}}>n_{\text{whitney}}$, known as Nash–Kuipe embedding theorem \citep{nash1954c1,kuiperc11}. G\"{u}nther finds a tighter lower bound on the dimensionality of the Euclidean space $n_{\text{nash}} \ge \max \{m(m+1)/2, m(m+3)/2 + 5 \}$ in the Nash embedding set-up \citep{gunther90}.

Although these theoretical results show the existence of such mappings, the remaining question is the feasibility of finding such mappings from an algorithmic perspective. \citep{baraniuk2009random,clarkson2008tighter} propose algorithms (with theoretical guarantees) based on the random projection methods to achieve an approximate isometric mapping. Also, some previous studies empirically show the feasibility of finding such mapping functions such as Isomap \citep{tenenbaum2000global}, Locally Linear Embedding \citep{roweis2000nonlinear}, Hessian Eigenmapping \citep{donoho2003hessian},  Laplacian Eigenmaps \citep{belkin2003laplacian}, Local Tangent Space Alignment \citep{zhang2004principal}, Local Fisher’s Discriminant Method \citep{sugiyama2007dimensionality}, Riemannian Manifold Learning \citep{lin2008riemannian}, and Local Manifold Learning-Based $k$-Nearest-Neighbor \citep{ma2010local} among many others.

The scope of our work remains in the Whitney embedding setup, which is less restrictive than Nash-Kuipe embedding that requires the isometry condition. Notice that, although the isometry property is an essential condition to preserve the geometrical structure, there is a trade-off between the dimensionality of the target Euclidean space and the isometry restriction, i.e., $n_{\text{nash}} > n_{\text{whitney}}$. Also, constructing a proper mapping function that preserves metrics between two spaces is more challenging and ongoing research.

\subsection{Other related works} \label{sec:otherworks}
\citet{courty2017learning} propose a method based on the siamese architecture \citep{chopra2005learning} to approximately embed the Wasserstein metric into the Euclidean metric. This method finds the backward map from the Euclidean metric to the initial Wasserstein metric in an autoencoder fashion \citep{hinton2006reducing} that is approximately invertible. A similar idea is used for point cloud embedding to speed-up Wasserstein distance calculations \citep{kawano2020learning}. \citet{xiao2018bourgan} embed sub-samples of datasets with an arbitrary metric into the Euclidean metric to address mode collapse for generative adversarial networks (GANs) \citep{goodfellow2014generative}. \citet{bramburger2021deep} aim to parametrize Poincar\'e map by a deep autoencoder architecture that enforces invertibility as a soft constraint.

\citet{otto2019linearly,lusch2018deep,gin2021deep} use deep autoencoder architecture to find a coordinate transformation that linearizes a nonlinear partial differential equation (PDE). \citep{lee2020model} use deep convolutional autoencoders to project a dynamical system onto a nonlinear manifold of fewer dimensions instead of the classical approach of finding a linear subspace. \citet{kim2021fast} construct a reduced-order model based on the autoencoder architecture with shallow multilayer perceptrons (MLPs) to find a proper linear subspace for a nonlinear PDE that describes advection-dominated phenomena.
To the best of our knowledge, the current work is the first paper that leverages the bijectivity of the invertible neural network to generate the desired embedding space for the data-driven paradigm.

\section{Model-free framework}\label{sec::model-free}
For completeness, we briefly summarize the distance-minimization model-free paradigm  \citep{kirchdoerfer2016data}. In this approach, the classical solver for a mechanics problem
(or generally speaking, a boundary value problem) 
 is reformulated as a minimization problem that finds an admissible solution satisfying the conservation laws with the minimum distance to a given database populated by discrete point clouds in the phase space. 
Due to its minimal model assumption about the constitutive behavior, it is known as a model-free method and
categorized as a non-parametric model.  
As shown in Fig. \ref{fig::overview}(a), the classical model-free paradigms often consist of two iterative algorithmic steps operated in the conservation (governing equation) and constitutive manifold respectively 
where the equilibrium states are first determined and projected onto the material data. 
In the current work, to simplify the formulation and avoid the usage of a nonlinear solver, the search of the equilibrium states is based on the energy norm originated from \citet{kirchdoerfer2016data}
whereas material data identification is conducted in the embedded Euclidean space of the constitutive manifold.

To obtain the equilibrium state, we consider pairs of strain-stresses $\{\vec{z}^*_a = (\eps^*_a, \sig^*_a) \}_{a=1}^{N_{\text{quad}}}$ at $N_{\text{quad}}$ quadrature points for a spatial domain discretized by finite elements. To find the equilibrium state that has the minimum distance to the given strain-stress 2-tuple, 
the following constrained optimization problem is solved, 

\begin{alignat*}{3}
    &\underset{\vec{z}}{\argmin} \int_{\Omega} d_{M}^2(\vec{z}, \vec{z}^*) \ d\Omega,
    \\
    &\text{subject to:} &&\diver{\sig} + \vec{b} = \vec{0} \quad \text{in} \ \Omega,
    \\
    & && \sig \cdot \vec{n} = \bar{\vec{t}} \quad \text{on} \ \Gamma_{\sig}, 
    \\
    & && \vec{u} = \bar{\vec{u}} \quad \text{on} \ \Gamma_{u}, 
    \\
    & && \eps = \frac{1}{2} \{ \grad{\vec{u}} + (\grad{\vec{u}})^T \} \quad \text{in} \ \Omega, 
    \label{eq:problemstatement}
\end{alignat*}
where $\vec{z} = (\eps, \sig)$, $\eps \subset \mathbb{R}^{m}$ is the strain tensor, $\sig \in \mathbb{R}^{m}$ is the stress tensor, $d_{M}(\cdot, \cdot)$ is the distance function, $\vec{b} \in \mathbb{R}^{N_{\text{dim}}}$ is the body force vector, $\vec{u} \in \mathbb{R}^{N_{\text{dim}}}$ is the displacement vector, $\vec{n}$ is the normal vector to the external boundary, $\bar{\vec{t}}$ is the prescribed traction on the boundary, $\bar{\vec{u}}$ is the prescribed displacement on the boundary, $\Omega \subset \mathbb{R}^{N_{\text{dim}}}$ is the volumetric domain, $\Gamma_{\sig} \subset \partial \Omega$ is the Dirichlet boundary surface, $\Gamma_{\sig} \subset \partial \Omega$ is the Neumann boundary surface, $N_{\text{dim}} = \{1, 2, 3\}$ is the space dimensionality, and $m = N_{\text{dim}} (N_{\text{dim}}+1)/2$ is the number of free components in the second-order symmetric tensor. The material state at each quadrature point $\vec{z}^*_a \subset \mathcal{D}$ where
$\mathcal{D}$ is the product space spanned by $N_{\text{data}}$ data points $\mathcal{D}= \overset{N_{\text{data}}}{\underset{b=1}{\bigtimes}} (\eps^*_b, \sig^*_b)$.
The search of the equilibrium state in the mechanical phase space $\mathcal{Z} \subset \mathbb{R}^{m} \times \mathbb{R}^{m}$ is conducted by minimizing the following norm $|\vec{z}|$:
\begin{equation}
    |\vec{z}|^2 = \frac{1}{2}\eps:\mathbb{C}:\eps +  \frac{1}{2}\sig:\mathbb{S}:\sig,
    \label{eq::energy-norm}
\end{equation}
where $\mathbb{C}$ and $\mathbb{S}$ are two fourth-order symmetric positive definite tensors. As suggested in \citep{kirchdoerfer2016data,he2020physics}, $\mathbb{S} = \mathbb{C}^{-1}$ is used herein. According to the introduced norm, the distance function reads as follows:
\begin{equation}
    d(\vec{z}, \vec{z}^*) = |\vec{z} - \vec{z}^*| = |(\eps-\eps^*, \sig-\sig^*)|
\end{equation}

The physical constraints listed in the problem statement are enforced via the Lagrange multipliers. Then, following the conventional calculus of variations procedure, we obtain the following Euler-Lagrange equations \citep{nguyen2020variational,he2020physics, bahmani2021kd}:
\begin{align}
    \mathcal{R}^u &= \int_{\Omega} \delta \vec{u} \cdot \frac{\partial \eps(\vec{u})}{\partial \vec{u}} : \mathbb{C} : (\eps - \eps^*) \ d \Omega = 0,
    \label{eq::disp-eq}
    \\
    \mathcal{R}^{\beta} &= -\int_{\Omega} \grad{\delta \vec{\beta}} : \sig d \Omega 
    + \int_{\Omega} \delta \vec{\beta} \cdot \vec{b} \ d\Gamma 
    + \int_{\Gamma_{\sig}} \delta \vec{\beta} \cdot \bar{\vec{t}} \ d\Gamma = 0,
    \label{eq::eta-eq}
    \\
    \mathcal{R}^{\sig} &= \int_{\Omega} \delta \sig : [\mathbb{S} : (\sig - \sig^*) - \grad{\vec{\beta}}] \ d\Omega = 0,
\end{align}
where $\vec{\beta} \in \mathbb{R}^{N_{\text{dim}}}$ is the Lagrange multiplier to enforce the balance of linear momentum, and $\delta$ indicates an arbitrary admissible variation, i.e., $\delta \vec{u} = \delta \vec{\beta} = \vec{0}$ on the Dirichlet boundary surface. Here, we reduce the number of independent fields by a local (point-wise) satisfaction of the last equation via:
\begin{equation}
    \sig = \sig^* + \mathbb{S}^{-1} : \grad{\vec{\beta}} \quad \text{in} \ \Omega.
    \label{eq::strs-eq}
\end{equation}

Given the material states $\{\vec{z}^*_a \}_{a=1}^{N_{\text{quad}}}$ where $\vec{z}^*_a \in \mathcal{D}$, the admissible mechanical states $\{\vec{z}_a \}_{a=1}^{N_{\text{quad}}}$ at each quadrature point can be obtained after solving the above linear, decoupled system of equations. This step can be considered, geometrically, as a global projection of material states to the conservation law manifold, which is called \textit{global} minimization or projection; see $\mathcal{P}^{L \mapsto G}$ in \fig \ref{fig::overview}(a). However, we do not know the optimal distribution of the material states \textit{a priori}. The distance-minimization method iterates between the space spanned by the data points and the conservation laws manifold until the optimal solution is found. After updating the mechanical states at each quadrature point, the new material states should be updated accordingly such that they have a minimum total distance to the mechanical states. This step is performed locally at each quadrature point to project the mechanical state onto the closest data point and is called \textit{local} minimization or projection; see $\mathcal{P}^{G \mapsto L}$ in \fig \ref{fig::overview}(a). The entire data-driven algorithm is provided in Algorithm \ref{algo::dd-solver}.

\begin{algorithm}[h!]
    \begin{algorithmic}[1]
    \State \textbf{Input:} Database $\{ (\eps^*_b, \sig^*_b) \}_{b=1}^{N_{\text{data}}}$, and numerical parameters $\mathbb{C}$, $\mathbb{S}$
    \State Randomly initialize the material state at quadrature points $\{ (\eps^*_a, \sig^*_a) \}_{a=1}^{N_{\text{quad}}}$ from the database
    \While {$\mathrm{\textit{not converged}}$}
        	\State Global projection by solving Eqs. \ref{eq::disp-eq} and \ref{eq::eta-eq}. \Comment{$\mathcal{P}^{L \mapsto G}$} 
        	\For{$a = 1:N_{\text{quad}}$} \Comment{local operations $\mathcal{P}^{G \mapsto L}$}
        	    \State Update mechanical state $(\eps_a, \sig_a)$ by Eq. \ref{eq::strs-eq}
        	    \State Update new material state $(\eps^*_b, \sig^*_b)$ by the local projection of $(\eps_a, \sig_a)$ via Eq. \ref{eq::local-old}
        	\EndFor
    \EndWhile
\caption{Data-driven solver}
\label{algo::dd-solver}
    \end{algorithmic}
\end{algorithm}

Here, we generalize the local minimization step of the original distance minimization method as follows:
\begin{equation}
    \underset{\vec{z^*} \in \mathcal{D}}{\argmin}  \quad d_{D}^2(\vec{z}_a, \vec{z}^*),
    \label{eq::local-old}
\end{equation}
where $d_{D}$ is an appropriate metric to measure the distance in the data space $\mathcal{D}$.
 In the original distance minimization setup (cf. \citet{kirchdoerfer2016data}), we have $d_{D} (\cdot, \cdot) \equiv d_{M}(\cdot, \cdot)$, and the local minimization step is the nearest neighbor search in the database to find the closest data point to a given mechanical state $\vec{z}_a$.
Here we hypothesize that an efficient metric to measure the distance in each manifold could be different because the conservation and constitutive manifolds may possess different geometrical structures.

\begin{figure}[h]
 \centering
 \subfigure[material data points]
{\includegraphics[width=0.4\textwidth]{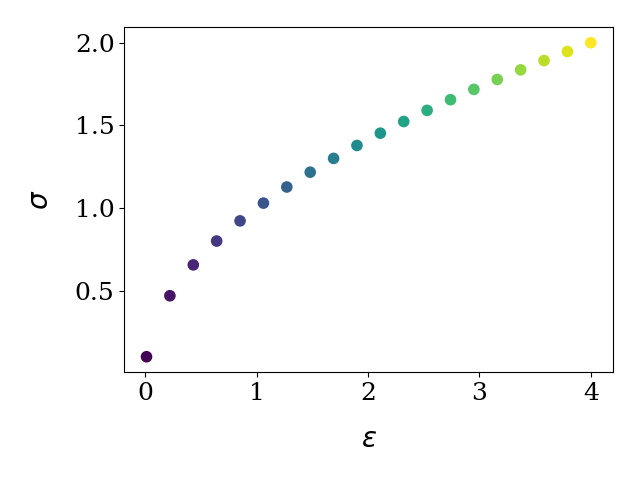}}
 \subfigure[mapped material data points]
{\includegraphics[width=0.4\textwidth]{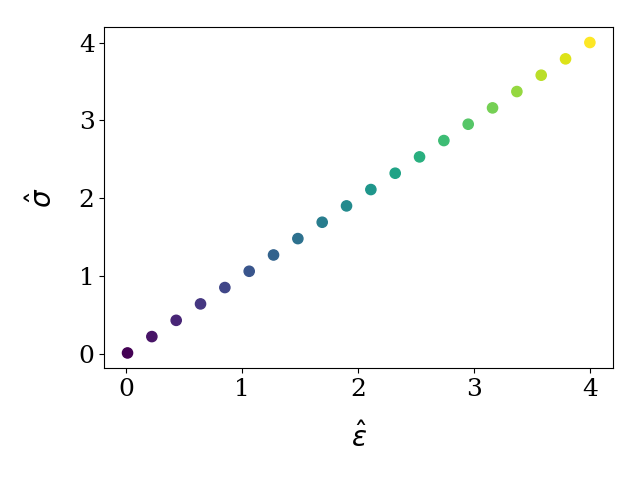}}
  \caption{(a) synthesized database by $\sigma = \sqrt{\epsilon}$ with 20 data points that are generated by the regular sampling along strain axis. (b) mapped database to a vector space by the invertible neural network. Colors show the data point number.
   \label{fig::1d-sqrt-data}}
\end{figure}

\subsection{Illustrative example 1: dependence of metrics for local projections}\label{sec::model-free-local-proj-ex}

\begin{figure}[h]
 \centering
 \subfigure[Nearest neighbor projection]
{\includegraphics[width=0.4\textwidth]{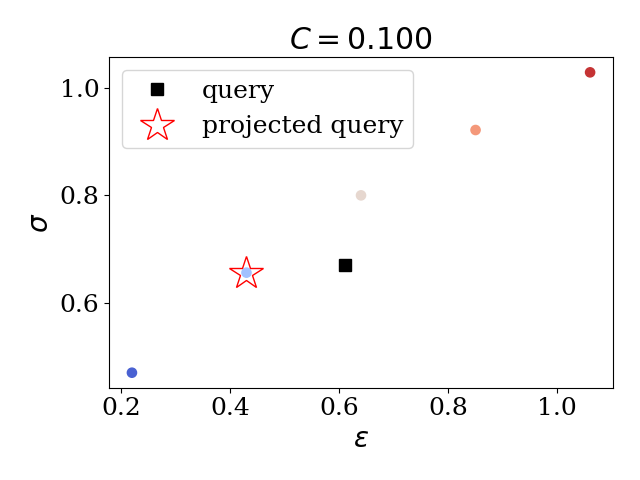}}
\hspace{0.01\textwidth}
 \subfigure[Nearest neighbor projection]
{\includegraphics[width=0.4\textwidth]{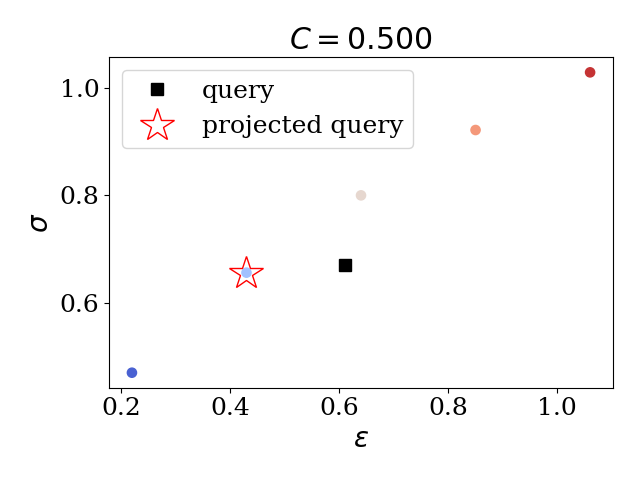}}
\hspace{0.01\textwidth}
 \subfigure[Nearest neighbor projection]
{\includegraphics[width=0.4\textwidth]{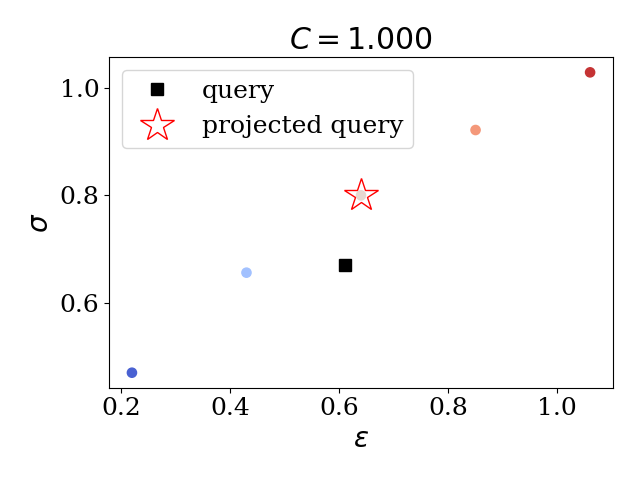}}
\hspace{0.01\textwidth}
 \subfigure[Manifold embedding projection (this paper)]
{\includegraphics[width=0.4\textwidth]{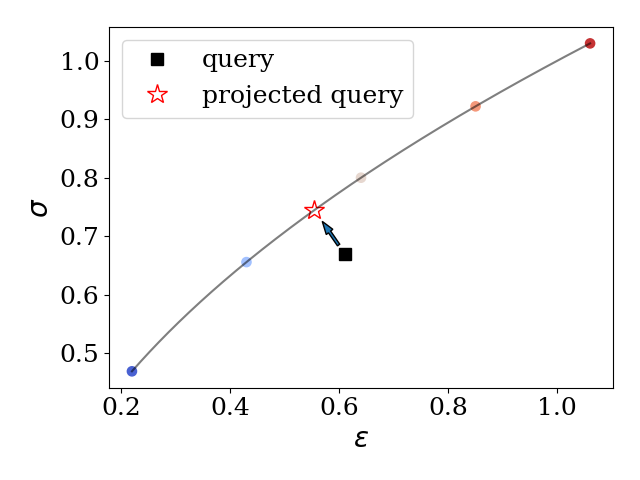}}
  \caption{ 
  Comparing local projections for one arbitrary query point $\vec{z} = (0.61, 0.67)$ shown by a black square marker between (a-c) the original distance minimization method and (d) manifold embedding method introduced in this paper.  A query point in the local minimization step is a point that belongs to the conservation manifold, found in the global minimization step, but not necessarily has the minimum distance to the material database. In (a-c), different $C$ parameter in Eq. \ref{eq::energy-norm} is used to show the importance of chosen norm. Points with colorful circular markers are material data points. The star marker shows the projected material state for the query point. The solid black curve in (d) indicates the underlying constitutive manifold used to synthesize the database.
   \label{fig::nns-C-eff-one-q}}
\end{figure}

To show the importance of choosing a suitable metric in the local minimization step, let's consider a simple one-dimensional problem where the material database is given in \fig \ref{fig::1d-sqrt-data}(a). \Fig \ref{fig::nns-C-eff-one-q}(a-c) studies the effect of parameter $C$ (see Eq. \ref{eq::energy-norm}) on the found closest data point $\vec{z}^*$ shown by red star marker to a query mechanical state $\vec{z} = (0.61, 0.67)$ shown by a black square marker. Notice that, in the one-dimensional setup, the forth-order tensor $\mathbb{C}$ becomes a scalar $C$. As we observe, based on the value of $C$, there exist two different responses. 

\begin{figure}[h]
 \centering
 \subfigure[Nearest neighbor projection]
{\includegraphics[width=0.4\textwidth]{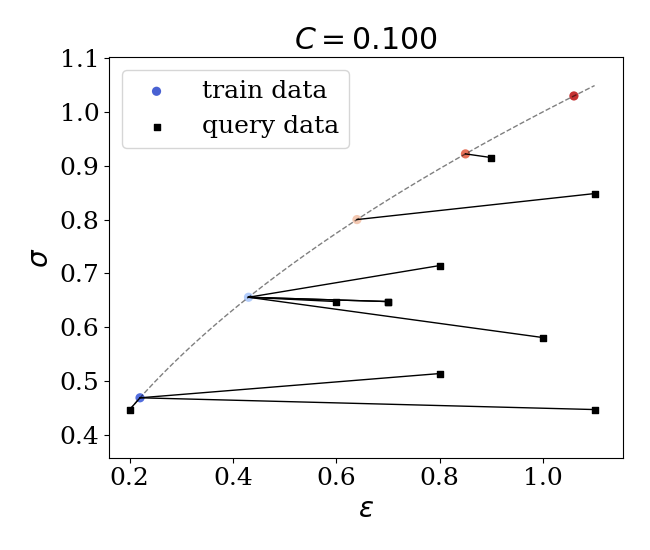}}
\hspace{0.01\textwidth}
 \subfigure[Nearest neighbor projection]
{\includegraphics[width=0.4\textwidth]{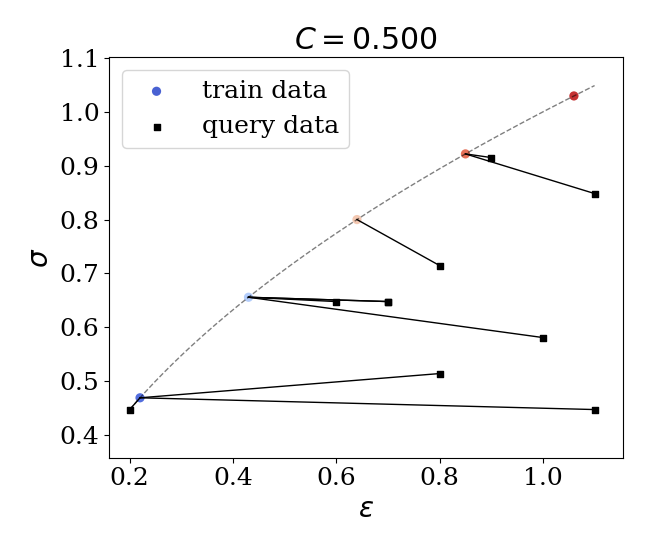}}
\hspace{0.01\textwidth}
 \subfigure[Nearest neighbor projection]
{\includegraphics[width=0.4\textwidth]{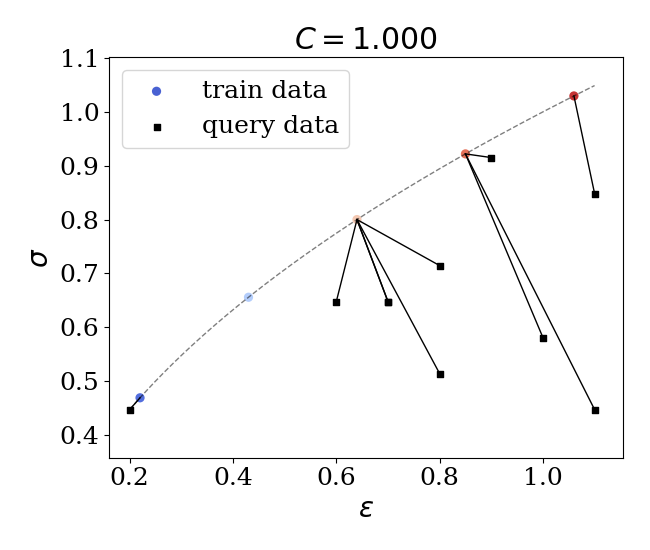}}
\hspace{0.01\textwidth}
 \subfigure[Manifold embedding projection (this paper)]
{\includegraphics[width=0.4\textwidth]{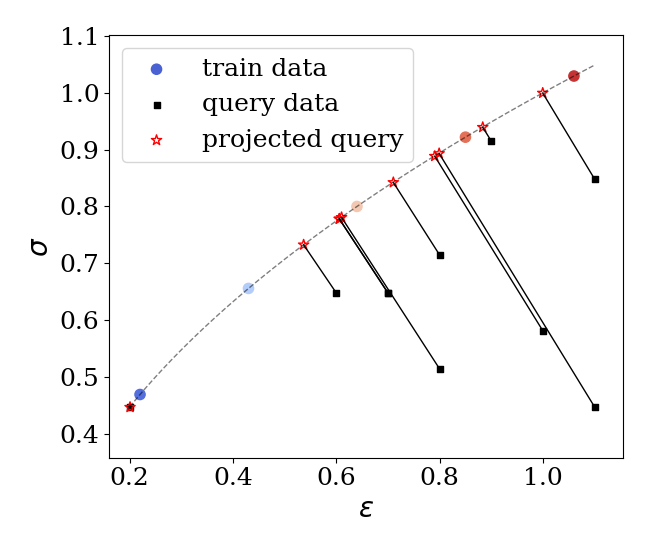}}
  \caption{Comparing local projections between (a-c) the original distance minimization method and (d) manifold embedding method introduced in this paper. Query points are shown by black square markers. In (a-c), different $C$ parameters in Eq. \ref{eq::energy-norm} are used to exhibit the dependence of projected results with respect to the equipped norm. The solid lines show the projection results. The dashed line indicates the underlying constitutive manifold used to synthesize the database (colorful dot points). The results in (a-c) suggest that the norm may considerably affect the final results, especially for those points too far away from the train data points.
   \label{fig::nns-C-eff}}
\end{figure}

To further investigate this issue, we consider 10 random query points shown by black square markers in \fig \ref{fig::nns-C-eff}(a-c). In this experiment, we observe that the final responses can vary among \textit{only} five different choices based on the different assigned values of the parameter $C$. This observation confirms that the choice of the norm may have a significant impact on the final results, which is also reported in \cite{leygue2018data}. Notice that increasing the amount of data might reduce the chance of encountering this spurious issue. Here, we aim to enhance the consistency of this local projection step by introducing an algorithm that implicitly explores the underlying geometry of the data space $\mathcal{D}$ to find the closest interpolated data point based on a \textit{data-driven} metric formulation.

\section{Revisited local minimization in constitutive manifold}\label{sec::new-local}
As demonstrated in, for instance, \citet{ibanez2017data} and \citet{ibanez2018manifold}, nonlinear constitutive data often appear to be belonging to a real Riemannian manifold.
While the distance between two very close points in the same tangential space of the manifold can be characterized 
by the Euclidean distance, the geodesics distance of data points in a constitutive manifold is generally different from the Euclidean counterpart.

Our major point of departure here is the introduction of an indirect measurement of distance for the local minimization problem. In this work, the distance to be minimized is neither measured by a norm of a single Euclidean space in \citet{kirchdoerfer2016data} nor that of a local linear patch formed by the nearest neighbors of a data point. Instead, the distance of the data points of an imaginary linear space where \textit{all} the data points 
lie on the nonlinear constitutive manifold are mapped onto. 
We consider a constitutive response as a nonlinear mapping that maps a $m$-dimensional input onto a $m$-dimensional output. The data of the material databases are instances of this constitutive response where each instance is stored as a $2m$-dimensional vector. For instance, 
the symmetric strain and stress tensors are stored as a 12-dimensional vector since both the strain and stress tensor in three-dimensional space consists of six independent components (degrees of freedom) due to their symmetries).

We propose to \textit{globally} transform the constitutive manifold into a new space that admits the vector space properties by a globally nonlinear mapping function $\mathcal{F}: (\mathcal{D} \subset \mathbb{R}^{2m}) \mapsto \mathbb{R}^{2m}$. Hence, if such a mapping function does exist and 
can be determined, the Euclidean norm is indeed a valid norm for the constructed space that embedded the constitutive manifold. 

An important property we want to achieve is to ensure that the image of $\mathcal{F}$ is a hyperplane. In particular, want to ensure that, for any data point $\hat{\vec{z}}$, the coordinates $ \hat{\vec{z}}_i$ and $ \hat{\vec{z}}_j$ can be linearly related by a
constant matrix $\hat{\tensor{K}}$, i.e., 
\begin{equation}
    \hat{\vec{z}}_j = \hat{\tensor{K}}_{ji} \hat{\vec{z}}_i \quad \text{for} \quad 1 \le i \le m, m < j \le 2m,
    \label{eq::linear-cond}
\end{equation}
where $\hat{\vec{z}} = \mathcal{F}(\vec{z})$. To justify this claim, the hyperplane parameterized by a constant non-zero normal vector $\hat{\vec{c}} \in \mathbb{R}^{2m}$ can be found straightforwardly such that $\hat{\vec{z}} \cdot \hat{\vec{c}} = 0$ and hence the image of the function $\mathcal{F}$ is a vector space embedded in $\mathbb{R}^{2m}$. In the rest of this subsection, we would assume such a mapping exists, and later we introduce an algorithm to find such a mapping through the expressivity power of neural networks \citep{hornik1989multilayer,balestriero2021learning}.

The image of $\mathcal{F}$ is a vector space only for points sampled from the data manifold $\vec{z}^* \subset \mathcal{D}$, therefore any point $\vec{z} \notin \mathcal{D}$ will not be mapped onto the constructed hyperplane, as schematically shown in \fig \ref{fig::overview}(b) where in Step 2 the mapped point $\vec{z}^{\#}$ does not belong to the hyperplane colored by blue. However, we can reasonably perform the distance minimization projection to project such a point onto the mapped data space with respect to the Euclidean metric. In addition, since the tangent space for the constructed hyperplane is globally constant everywhere on the hyperplane, we can relax the discrete minimization statement in Eq. \ref{eq::local-old} to its continuous version in the mapped space. This continuous relaxation changes the initial NP-hard problem \citep{bahmani2021kd} to a tractable one. We provide the details of this new local projection scheme as follows. 

For the ease of interpretation and explanation, we call the first $m$ components of the mapped variable $\hat{\vec{z}}$ pseudo-strain $\hat{\eps}$ and the rest pseudo-stress $\hat{\sig}$, i.e.,
$\hat{\vec{z}} = \begin{bmatrix}
\begin{tabular}{ c}
$\hat{\eps}$\\
\hline
$\hat{\sig}$
\end{tabular}
\end{bmatrix}$.
To find a closed-form solution for the local minimization projection, we restrict the $\hat{\tensor{K}}$ to be symmetric positive-definite in Eq. \ref{eq::linear-cond}, i.e., $\hat{\tensor{K}} \in \text{SPD}(m)$. We introduce the local minimization in the mapped space as follow:
\begin{align*}
    &\underset{\hat{\vec{z}} \in \mathbb{R}^{2m}}{\argmin}  \quad \hat{d}^2(\vec{z}^{\#}, \hat{\vec{z}}),
    \\
    &\text{subject to:} \quad \hat{\sig} = \hat{\tensor{K}} \hat{\eps},
\end{align*}
where $\hat{d}(\cdot, \cdot)$ is the energy distance in the mapped domain as follows:
\begin{equation}
    \hat{d}^2(\vec{z}^{\#}, \hat{\vec{z}}) = 
    \frac{1}{2} (\eps^{\#} - \hat{\eps})^T \hat{\tensor{K}} (\eps^{\#} - \hat{\eps})
    +
    \frac{1}{2} (\sig^{\#} - \hat{\sig})^T \hat{\tensor{K}}^{-1} (\sig^{\#} - \hat{\sig}).
    \label{eq::local-metric}
\end{equation}
We obtain the following unique solution for the above quadratic objective by setting the objective's derivative to zero and using basic linear algebra manipulations:
\begin{align}
    &\hat{\eps} = \frac{1}{2}(\eps^{\#} + \hat{\tensor{K}}^{-1}\sig^{\#}),
    \label{eq::local-dist-min-eps}
    \\
    &\hat{\sig} = \frac{1}{2}(\sig^{\#} + \hat{\tensor{K}}\eps^{\#}),
    \label{eq::local-dist-min-sig}
    \\
    &\hat{\vec{z}}^T = \mathbb{P}^T(\vec{z}^{\#}) = [\hat{\eps}^T, \hat{\sig}^T],
\end{align}
where $\mathbb{P}$ is the projection operator to find the closest point on the hyperplane, see step $2\to3$ in \fig \ref{fig::overview}(b).

Next, we should find the corresponding point in the actual data space since the governing equations are established with respect to the actual data space. Therefore, in our introduced framework, a proper map function $\mathcal{F}$ should be \textit{globally} bijective (invertible). In the next section, we introduce a method to find an appropriate bijective map function parameterized by neural networks.

\remark{The energy metric in Eq. \ref{eq::local-metric} is equivalent to the Euclidean metric and they can be mapped bijectively upon a linear transformation \citep{bahmani2021kd}. This metric can be reduced into the Euclidean metric by setting $\hat{\tensor{K}} = \tensor{I}_m$ where $\tensor{I}_m$ is the $m\times m$ identity matrix.}\label{rk::l2-equivl}

\section{Neural network embedding of constitutive manifold}\label{sec::nn-embd}

\begin{figure}[h]
 \centering
{\includegraphics[width=0.6\textwidth]{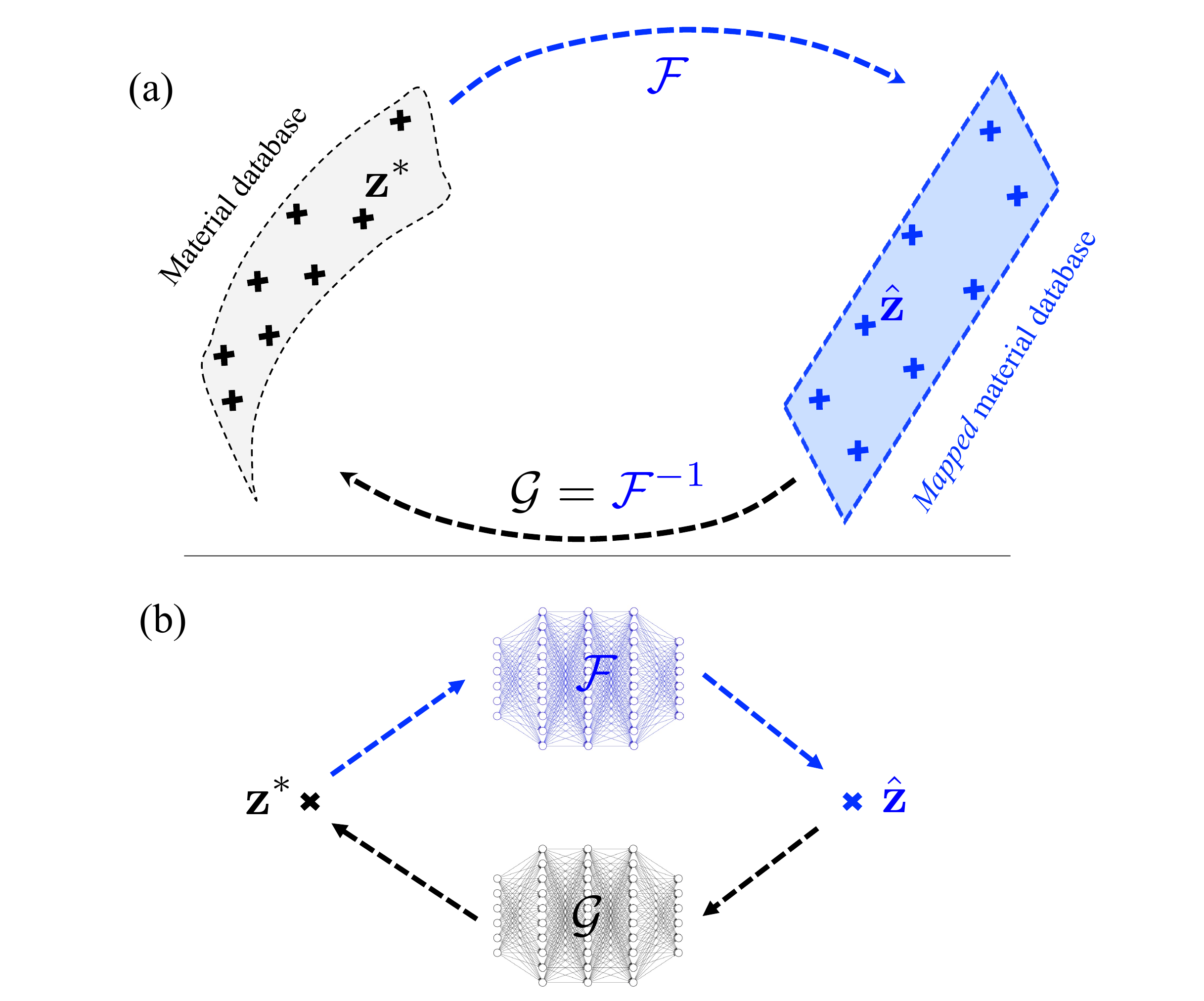}}
  \caption{(a) \textit{global} mapping of the material data space into a vector space by a bijective function and vice-versa. (b) a vanilla autoencoder architecture to parametrize a bijective function by encoder $\mathcal{F}$ and decoder $\mathcal{G}$ deep neural networks.
   \label{fig::nn-map}}
\end{figure}

 We are interested in finding an appropriate function $\mathcal{F}$ with the desired properties mentioned in the previous section to map the data points of the actual database onto a hyperplane (a Euclidean space), see \fig \ref{fig::nn-map}(a). We hypothesize that such a mapping function can be found in the class of multilayer perceptrons (MLPs) due to their expressiveness power \cite{hornik1989multilayer}. In a general setup, MLPs are not guaranteed to be invertible.
However, we may use the similar idea introduced in the autoencoder architecture
 (cf. \citet{hinton2006reducing,bengio2007scaling,vincent2008extracting,baldi2012autoencoders}) to define different MLPs for the forward $\mathcal{F}$ and backward $\mathcal{G}$ functions, see \fig \ref{fig::nn-map}(b). The connection between forward and backward maps, i.e., $\mathcal{G}=\mathcal{F}^{-1}$, is incorporated as a soft constraint in the optimization statement. Therefore, our proposed local projection can be expressed as the following optimization problem:
\begin{equation}
    \underset{
    \vec{\theta} \in \mathbb{R}^{N_{\mathcal{F}}}, \vec{\beta} \in \mathbb{R}^{N_{\mathcal{G}}}, \hat{\tensor{K}}\in \text{SPD}(m)
    }
    {\argmin}
    \sum_{i=1}^{N_{\text{data}}}
    ||\hat{\sig}_i - \hat{\tensor{K}} \hat{\eps}_i||_2^2
    +
    ||\vec{z}_i^* - \mathcal{G}(\mathcal{F}(\vec{z}_i^*, \vec{\theta}), \vec{\beta})||_2^2,
    \label{eq::loss-auto-enc}
\end{equation}
where functions $\mathcal{F}$ and $\mathcal{G}$ are MLPs parameterized by unknown vectors $\vec{\theta}$ and $\vec{\beta}$ with sizes $N_{\mathcal{F}}$ and $N_{\mathcal{G}}$, respectively. In this notation, these vectors concatenate weights and biases. The first term in the above objective is designated to enforce the linearity of the target vector space. The second term enforces the bijectivity constraint also known as the reconstruction error for the autoencoder architecture where $\mathcal{F}$ and $\mathcal{G}$ are encoder and decoder functions. The above architecture is a naive formulation of the problem statement explained in the previous section, and we call it the \textit{vanilla} autoencoder formulation.

By leveraging the expressive power of the deep neural networks \citep{hornik1989multilayer, raghu2017expressive}, 
we argue that, as long as (1) the training is successful and (2) the unsampled data not used in the training process 
are indeed on the same manifold of the training data, the numerical value of 
$\hat{\tensor{K}}_{\text{opt}}$ in \ref{eq::loss-auto-enc} is not consequential. Since 
it is equivalent to any \textit{fixed} $\hat{\tensor{K}}_{\text{fix}} \in \text{SPD}(m)$ up to a linear transformation  $\hat{\tensor{K}}_{\text{opt}} = \tensor{T} \hat{\tensor{K}}_{\text{fix}}; \tensor{T} \in \text{SPD}(m)$, every two members of SPD group commutes by the matrix multiplication. This linear transformation can be indirectly found as the outcome of a successfully trained neural network in which we are interested in the properties of the embedding space but not 
its explicit value. Therefore, in this work, we fix the matrix $\hat{\tensor{K}}_{\text{fix}} \in \text{SPD}(m)$ to simplify the optimization problem.

 The optimization over MLPs is known to be non-convex \citep{goodfellow2016deep}; hence, finding the global minimizer of the above objective is not trivial in a general setting. Even for the near-optimal local minimizers, the bijectivity constraint cannot be achieved precisely. Moreover, this optimization problem is \textit{multi-objective}, which makes it necessary to consider the possibility of conflict situations where  Pareto efficiency could be a primary concern \citep{yu2020gradient,bahmani2021training}. To address these drawbacks, we incorporate a new architecture in the next subsection that leads to neural network predictions preserving the bijectivity condition by construction and hence bypassing the need to handle conflicting objectives. 

\subsection{Invertible neural network}

\begin{figure}[h]
 \centering
{\includegraphics[width=0.6\textwidth]{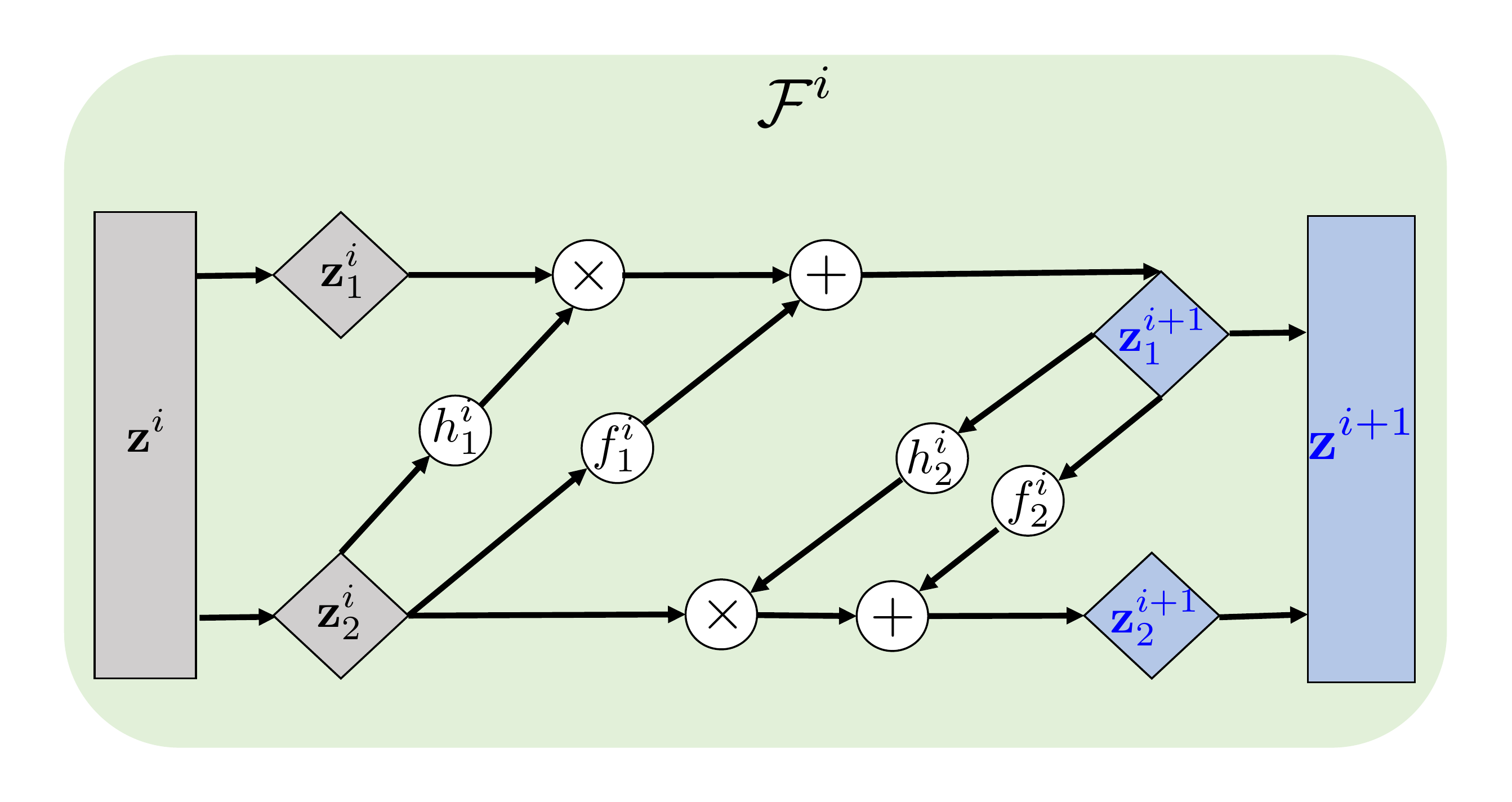}}
  \caption{Computational graph of the $i^{\text{th}}$ coupling layer in the invertible architecture. $h_1^i$, $h_2^i$, $f_1^i$, and $f_2^i$ are any arbitrary functions such as MLP. $\vec{z}^i$ and $\vec{z}^{i+1}$ are the input and output feature vectors of the coupling layer. 
   \label{fig::inv-layer}}
\end{figure}

Due to the drawbacks of the vanilla autoencoder framework mentioned in the previous section, we formulate a new machine learning strategy built upon a type of neural network that is \textit{bijective} by its construction (as an inductive bias). The invertible neural network is built upon several coupling layers. A coupling layer maps the $i^{\text{th}}$ layer's feature vector $\vec{z}^i \in \mathbb{R}^{n_i}$ to the $(i+1)^{\text{th}}$ layer's feature vector $\vec{z}^{i+1} \in \mathbb{R}^{n_{i+1}}$  via the function $\mathcal{F}^i:\vec{z}^i \mapsto \textcolor{blue}{\vec{z}^{i+1}}$ customized as follows \citep{dinh2014nice,dinh2016density,ardizzone2018analyzing,beitler2021pie}:
\begin{align}
    &\textcolor{blue}{\vec{z}_1^{i+1}} = \vec{z}_1^i \odot h^i_1(\vec{z}_2^i) + f^i_1(\vec{z}_2^i),
    \label{eq::forward-inv-net-1}
    \\
    &\textcolor{blue}{\vec{z}_2^{i+1}} = \vec{z}_2^i \odot h^i_2(\textcolor{blue}{\vec{z}_1^{i+1}}) + f^i_2(\textcolor{blue}{\vec{z}_1^{i+1}}),
    \label{eq::forward-inv-net-2}
\end{align}

where $\odot$ is the Hadamard product (element-wise multiplication), and $h^i_1$, $h^i_2$, $f^i_1$, and $f^i_2$ are  four arbitrary functions called internal functions. 
In each coupling layer, before the forward pass, the input feature $\vec{z}^i$ is divided into two arbitrary, disjoint halves $\vec{z}^i_1 \in \mathbb{R}^{m_1^i}$ and $\vec{z}^i_2 \in \mathbb{R}^{m_2^i}$ where $m_1^i + m_2^i = n_i$. Notice that the internal functions can be deep neural networks to increase the complexity and expressivity of each layer. In this work, we parametrize these functions via MLPs.

Because of the simple yet elegant structure in each coupling layer, the backward map $(\mathcal{F}^i)^{-1}: \textcolor{blue}{\vec{z}^{i+1}} \mapsto \vec{z}^i$ is readily available as follows: 
\begin{align}
    \vec{z}_2^i &= (\textcolor{blue}{\vec{z}_2^{i+1}} - f^i_2(\textcolor{blue}{\vec{z}_1^{i+1}}))
    \odot
    \frac{1}{h^i_2(\textcolor{blue}{\vec{z}_1^{i+1}})}
    \label{eq::backward-inv-net-1}
    \\
    \vec{z}_1^i &= (\textcolor{blue}{\vec{z}_1^{i+1}} - f^i_1(\vec{z}_2^i))
    \odot
    \frac{1}{h^i_1(\vec{z}_2^i)}.
    \label{eq::backward-inv-net-2}
\end{align}
A deep bijective, invertible neural network $\mathcal{F}: (\vec{z}_{\text{in}} \in {\mathbb{R}^{n_{\text{in}}}}) \mapsto (\vec{z}_{\text{out}} \in {\mathbb{R}^{n_{\text{out}}}})$ can be constructed by the composition of $L$ coupling layers:
\begin{equation}
    \mathcal{F}(\vec{z}; \vec{\theta}) = \mathcal{F}^{L} \circ \mathcal{F}^{L-1} \circ \cdots \circ \mathcal{F}^1 (\vec{z}).
\end{equation}
where $\vec{\theta}$ concatenates all the parameters defined for deep neural networks $\{h_1^i, h_2^i, f_1^i, f_2^i\}_{i=1}^L$.

We tailor the introduced architecture for our purpose as follows. The input feature vector $ \vec{z}^1 \leftarrow \vec{z}^* \in \mathcal{R}^{2m}$ can be naturally halved into $\vec{z}^1_1 \leftarrow \eps^*$ and $\vec{z}^1_2 \leftarrow \sig^*$. In our experience, we find that a single coupling layer is sufficient for the problems solved herein. In this setup, the output feature vector becomes $(\vec{z}^2)^T = [(\vec{z}_1^2)^T, (\vec{z}_2^2)^T]$ and $\bar{\eps} \leftarrow \vec{z}^2_1$ and $\bar{\sig} \leftarrow \vec{z}^2_2 $. Also, to reduce the training computational cost, we set $h_1^1 \equiv h_2^1 \equiv \vec{J}_m$ and $f_1^1 = \vec{0}_m$ where $\vec{J}_m$ and $\vec{0}_m$ are vectors of size $m$ filled by ones and zeros, respectively. This architecture is equivalent to the network introduced in \citep{dinh2014nice}.

Utilizing this architecture, the previous multi-objective loss function in Eq. \ref{eq::loss-auto-enc} becomes a single objective optimization problem as follows:
\begin{equation}
    \underset{
    \vec{\theta} \in \mathbb{R}^{N_{\mathcal{F}}}
    }
    {\argmin}
    \sum_{i=1}^{N_{\text{data}}}
    ||\hat{\sig}_i - \hat{\tensor{K}}_{\text{fix}} \hat{\eps}_i||_2^2,
    \label{eq::loss-inv-net}
\end{equation}
where $\hat{\vec{z}} = \mathcal{F}(\vec{z}^*; \vec{\theta})$. In the following subsection, we showcase the application of the introduced local projection via a simple example.

\begin{algorithm}[h!]
    \begin{algorithmic}[1]
    \State \textbf{Input: a query point $(\eps, \sig)$} 
    \State Map forward $(\eps, \sig)$ via Eqs. \ref{eq::forward-inv-net-1} and \ref{eq::forward-inv-net-2} to $(\eps^{\#}, \sig^{\#})$.
    \State Find the closest point $(\hat{\eps}, \hat{\sig})$ on the constructed vector space to $(\eps^{\#}, \sig^{\#})$ by Eqs. \ref{eq::local-dist-min-eps} and \ref{eq::local-dist-min-sig}.
    \State Map backward $(\hat{\eps}, \hat{\sig})$ via the inverse operations in Eqs. \ref{eq::backward-inv-net-1} and \ref{eq::backward-inv-net-2} to $(\eps^*, \sig^*)$.
\caption{Query inference with the introduced local projection}
\label{algo::local-proj-inv}
    \end{algorithmic}
\end{algorithm}

\remark{In all of the examples solved in this paper, we set $\hat{\tensor{K}}_{\text{fix}}$ equal to the identity tensor which is corresponding to the regular l2-norm projection as discussed in Remark \ref{rk::l2-equivl}.}

\remark{Our proposed mapping operation $\mathcal{F}$ is an embedding (not immersion) operation since it is globally invertible and homeomorphism to its image.}

\remark{The introduced method in this paper relaxes the discrete optimization step of the original distance minimization to a continuous local projection onto a constructed smooth hyperplane; hence it breaks the NP-hardness.}

\subsection{Illustrative examples 2}
Here, we walk through the details of the introduced local projection method via the simple 1D example used in Section \ref{sec::model-free-local-proj-ex}. 

\begin{figure}[h]
 \centering
{\includegraphics[width=0.5\textwidth]{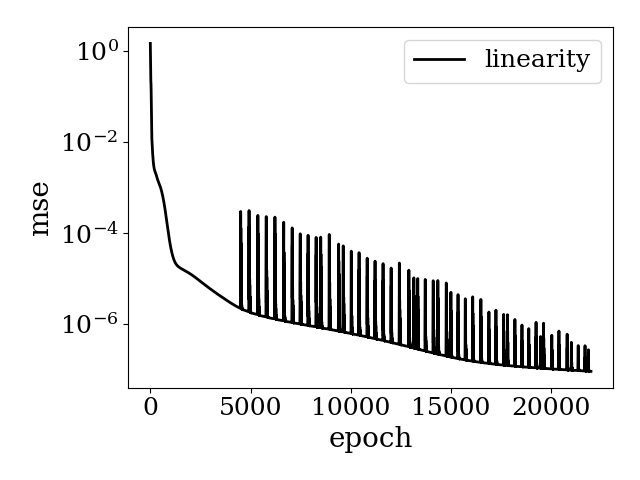}}
  \caption{Training performance of the invertible neural network used for the database shown in \fig \ref{fig::1d-sqrt-data}(a).
   \label{fig::sqrt-learning}}
\end{figure}

Recall that we construct the embedded vector space of the constitutive manifold in an offline manner based on the data points in an available database. The database we use is similar to the previous example, see \fig \ref{fig::1d-sqrt-data}(a). We utilize one invertible coupling layer with the internal function parameterized by an \texttt{elu} MLP \citep{clevert2015fast}. This MLP has three hidden layers. Each layer consists of 5 units initialized by the uniform Kaiming approach \citep{he2015delving}. The objective is minimized by the Adam method \citep{kingma2014adam} with the initial learning rate of 5e-3 in a full-batch manner. To enhance the training step, we use \texttt{ReduceLROnPlateau} learning scheduler of \texttt{PyTorch} library \cite{paszke2019pytorch} to reduce the learning rate by the factor of 0.91 every 50 iterations after the first 2000 iterations. The minimum learning rate is set to 1e-6. These hyperparameters are tuned manually with satisfactory results as shown in \ref{fig::sqrt-learning}. 
A more rigorous hyperparameter tunning might, to some degrees, further improve the performance, but such an endeavor is not the focus of this study \citep{bardenet2013collaborative, fuchs2021dnn2, heider2021offline}.

\Fig \ref{fig::sqrt-learning} summarizes the training performance. As this result suggest, the invertible architecture finds a near-optimal mapping function that can approximately satisfy the introduced loss function in Eq. \ref{eq::loss-inv-net} with the error in the order of $O(-6)$. For a better comparison, the mapped database in the constructed vector space is plotted in \fig \ref{fig::1d-sqrt-data}(b). The linearity of the 
data expressed in the coordinates of the embedded space is obvious. 

\begin{figure}[h]
 \centering
 \subfigure[regular grid points in strain-stress space]
{\includegraphics[width=0.4\textwidth]{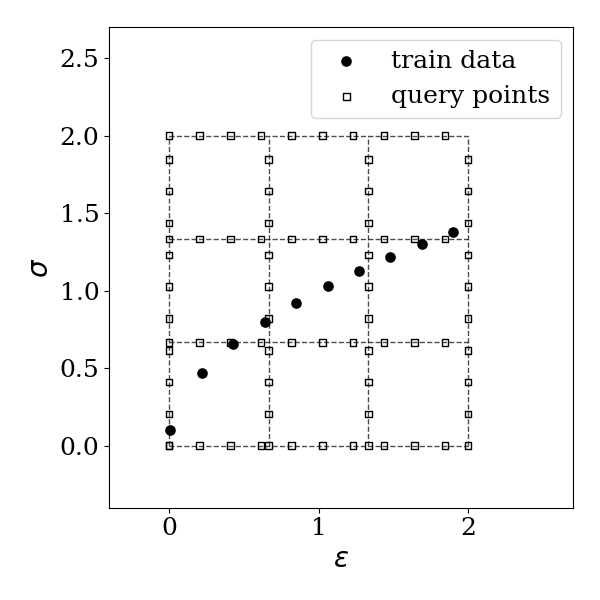}}
\hspace{0.01\textwidth}
 \subfigure[grid deformation after mapping]
{\includegraphics[width=0.4\textwidth]{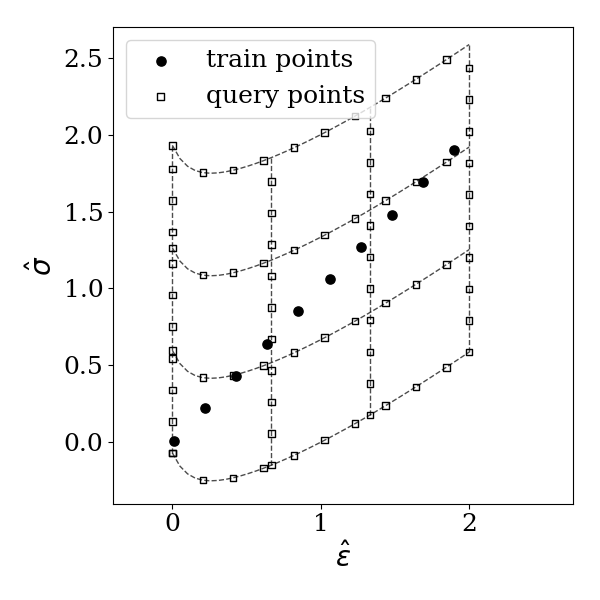}}
  \caption{ (a) a regular grid in the strain-stress space, (b) the deformation of this grid in the mapped domain.
   \label{fig::domain-distortion-sqr-data}}
\end{figure}

As discussed earlier, if a query point does not belong to the data manifold, it will be out of the constructed hyperplane. Hence, we define a Euclidean projection to project the mapped query point onto the constructed hyperplane, see Eqs. \ref{eq::local-dist-min-eps} and \ref{eq::local-dist-min-sig}. To show this and depict the effect of the mapping function, we present the deformation of a regular grid in the strain-stress space caused by the mapping function $\mathcal{F}$  in \fig \ref{fig::domain-distortion-sqr-data}. The mapping function can be thought as a  deformation of  the domain of the actual data points (circular markers in \fig \ref{fig::domain-distortion-sqr-data}(a)) such that the resultant deformed configuration becomes a hyperplane, i.e., a straight line in 1D setup; see  circular markers in \fig \ref{fig::domain-distortion-sqr-data}(b). However,  points out of the data manifold, i.e., square markers in \ref{fig::domain-distortion-sqr-data}(a), will not be placed on the constructed hyperplane.

\begin{figure}[h]
 \centering
 \subfigure[step 0: requested query]
{\includegraphics[width=0.4\textwidth]{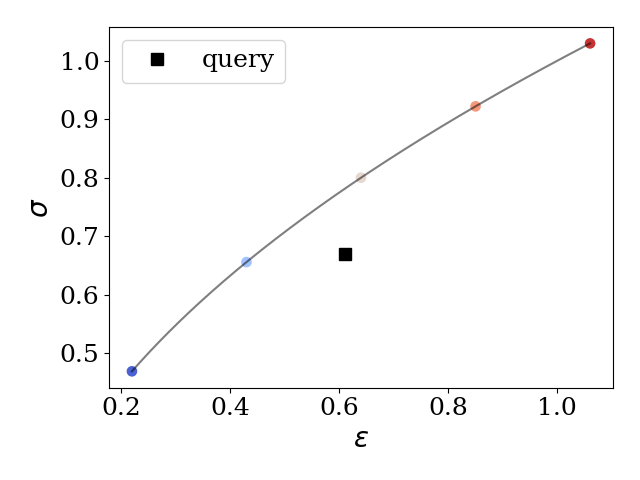}}
\hspace{0.01\textwidth}
 \subfigure[step 1: forward map to a vector space]
{\includegraphics[width=0.4\textwidth]{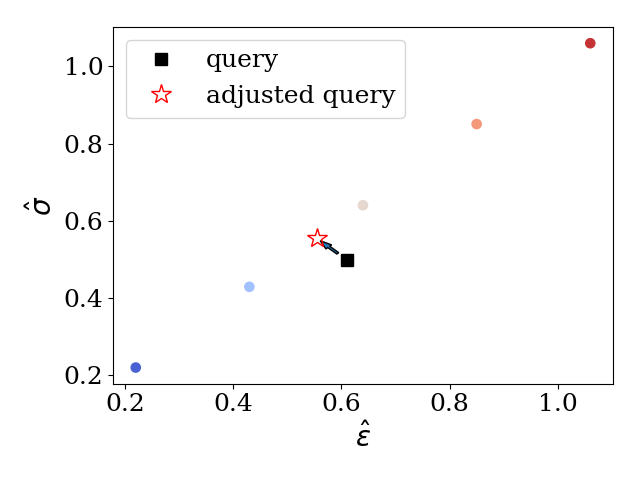}}
\hspace{0.01\textwidth}
 \subfigure[step 2: backward map to the real space]
{\includegraphics[width=0.4\textwidth]{figure/one_query_amb_sol.png}}
  \caption{ Summary of the steps for the introduced local projection: (a) a query point shown by a black square marker is requested to be projected onto the database. (b) the trained neural network is first applied to embed the database and the query point to the space that has vector space properties for database points. Moreover, in this step, the mapped query point should be adjusted to the constructed vector space for the database, see red star point in (b). (c) the adjusted point in (b) is finally mapped back to the real data space via the inverse operation. The black curve line shows the underlying material manifold. Colored circle points belong to the database.
   \label{fig::proj-method-for-one-q}}
\end{figure}

We show the steps of our local projection method at the query time in \fig \ref{fig::proj-method-for-one-q}, see Algorithm \ref{algo::local-proj-inv}.
In \fig \ref{fig::proj-method-for-one-q}(a) the similar mechanical state (depicted by a black square marker) to the problem in \fig \ref{fig::nns-C-eff-one-q} is chosen to be projected onto the data space. First, we map the query point, using Eqs. \ref{eq::forward-inv-net-1} and \ref{eq::forward-inv-net-2}, onto the constructed vector space by the trained invertible neural network, see the black square marker in \fig \ref{fig::proj-method-for-one-q}(b). Then, we find the closest point on the vector space to the mapped query point by Eqs. \ref{eq::local-dist-min-eps} and \ref{eq::local-dist-min-sig}, see the red star marker in \fig \ref{fig::proj-method-for-one-q}(b). Finally, we map back to the actual data space via the inverse operations Eqs. \ref{eq::backward-inv-net-1} and \ref{eq::backward-inv-net-2}, see red star marker in \fig \ref{fig::proj-method-for-one-q}(c). As the final result suggests, the proposed local projection can find the closest material point in the interpolation region of the data space and admits the underlying manifold. Notice that, even qualitatively, the found closest point is closer than the found points by the nearest neighbour approach, c.f. \fig \ref{fig::nns-C-eff-one-q}, and still the found point belongs to the underlying data generator manifold (black curve line in \fig \ref{fig::proj-method-for-one-q}).

For a better comparison with the nearest neighbor method, we apply our introduced local projection method to the similar query points used in \fig \ref{fig::nns-C-eff}(a-c). The result is indicated in \fig \ref{fig::nns-C-eff}(d) where the found local projections by our method provide a smooth transition from one point to the other compared with \fig \ref{fig::nns-C-eff}(a-c) where there were only five possible discrete choices. Our scheme can accurately interpolate between data points such that the projected values belong to the underlying data manifold.

\subsection{Illustrative examples 2: a sanity check through the
lens of convex interpolation on the manifold}

\begin{figure}[h]
 \centering
 \subfigure[$\alpha = 0.2$]
{\includegraphics[width=0.4\textwidth]{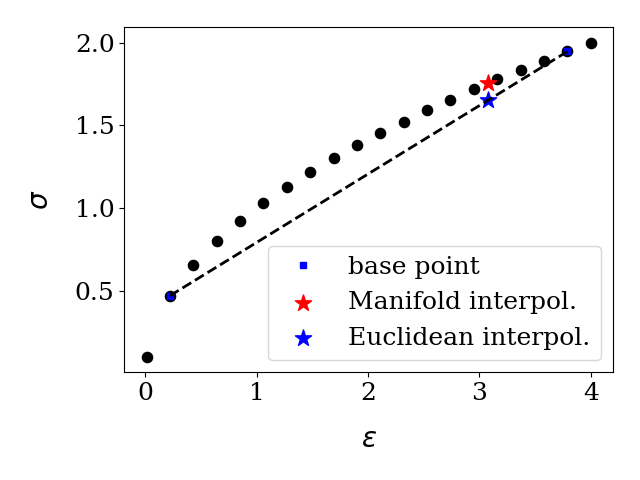}}
\hspace{0.01\textwidth}
 \subfigure[$\alpha = 0.4$]
{\includegraphics[width=0.4\textwidth]{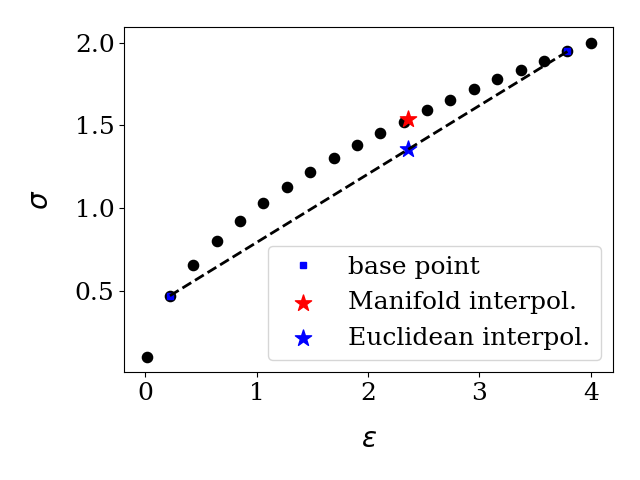}}
\hspace{0.01\textwidth}
 \subfigure[$\alpha = 0.6$]
{\includegraphics[width=0.4\textwidth]{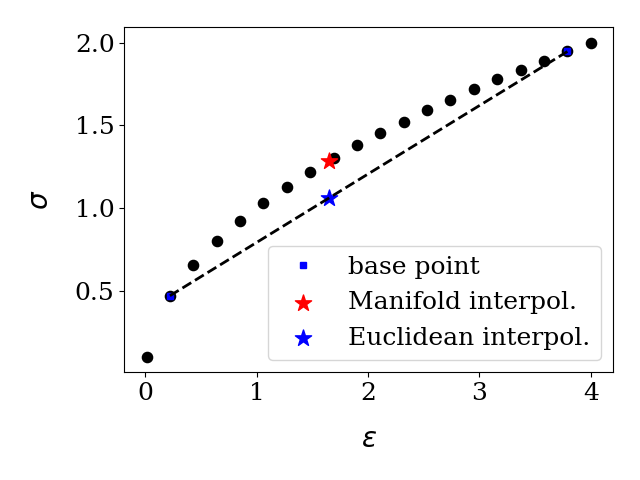}}
\hspace{0.01\textwidth}
 \subfigure[$\alpha = 0.8$]
{\includegraphics[width=0.4\textwidth]{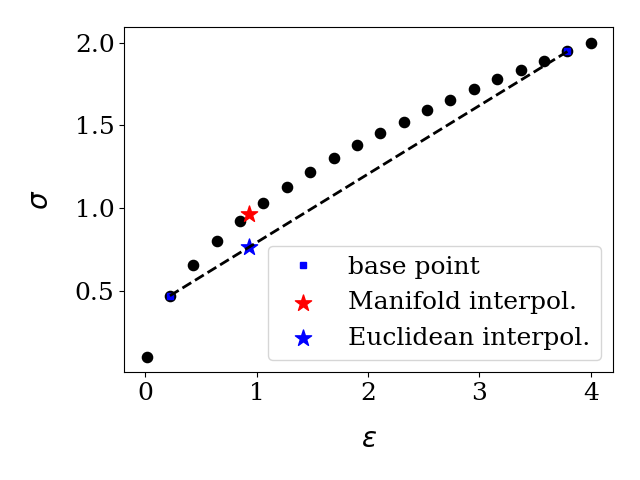}}
\hspace{0.01\textwidth}
  \caption{Interpolation capability in the regular data space and the constructed vector space for different values of $\alpha$ defined in Eq. \ref{eq::conx-intrpl}. Interpolated points (red star markers) in the constructed vector space are on the underlying material manifold. However, not surprisingly, the interpolation in the actual data space results in out of manifold points (blue star markers).
   \label{fig::conx-intpl}}
\end{figure}

Here, we examine the linearity of the constructed vector space by its convex interpolation capability. We claim that if the constructed space is truly vector space and the mapping function truly respects the underlying manifold, then any convex interpolation between two arbitrary points on the constructed space should result in a point on the actual manifold after the backward map. A convex interpolation between two points $\hat{\vec{z}}_s$ and $\hat{\vec{z}}_e$ is defined as follows:
\begin{equation}
    \hat{\vec{z}}_{\alpha} = \alpha \hat{\vec{z}}_s + (1 - \alpha) \hat{\vec{z}}_e; \quad \alpha \in (0, 1).
\label{eq::conx-intrpl}
\end{equation}

We examine this property in \fig \ref{fig::conx-intpl}. Start and end points $\hat{\vec{z}}_s$ and $\hat{\vec{z}}_e$ are shown by blue square markers. A red star marker depicts the interpolated point in the constructed vector space. The interpolated point in the actual data space is depicted as a blue star marker. As this experiment shows, the interpolation in the constructed vector space is equivalent to the interpolation over the underlying manifold. However, the regular interpolation in the actual data space results in points out of the underlying data manifold.

\section{Numerical Examples} \label{sec::numericalexample}
In this section, we benchmark four numerical problems to examine the accuracy and robustness of the proposed scheme in various scenarios. Importantly, we compare the efficiency of this method with the original distance minimization scheme in the limited data regime. First, in Section \ref{subsec::1D-bar}, we solve a 1D bar problem for two databases with different availability of data. Moreover, we compare the robustness of the proposed invertible network with its vanilla autoencoder variant. Second, we use the same material database to solve a 3D truss system under loading-unloading conditions in Section \ref{subsec::truss}. In this problem, we point to a circumstance where the original distance minimization may predict a spurious dissipation mechanism for an elastic material. Then, in Section \ref{subsec::heat}, we showcase the application of the proposed method for a heat conduction problem which is studied in the literature as well. In the last example shown in Section \ref{subsec::2D-hole}, we compare the efficiency of both methods for a plane strain problem that possesses stress concentration due to the geometrical imperfection. All the problems studied in this work deal with nonlinear material behavior. The databases will be publicly available for third-party validation exercises (upon acceptance of this manuscript).

\subsection{Nonlinear 1D bar}\label{subsec::1D-bar}

\begin{figure}[h]
 \centering
{\includegraphics[width=0.3\textwidth]{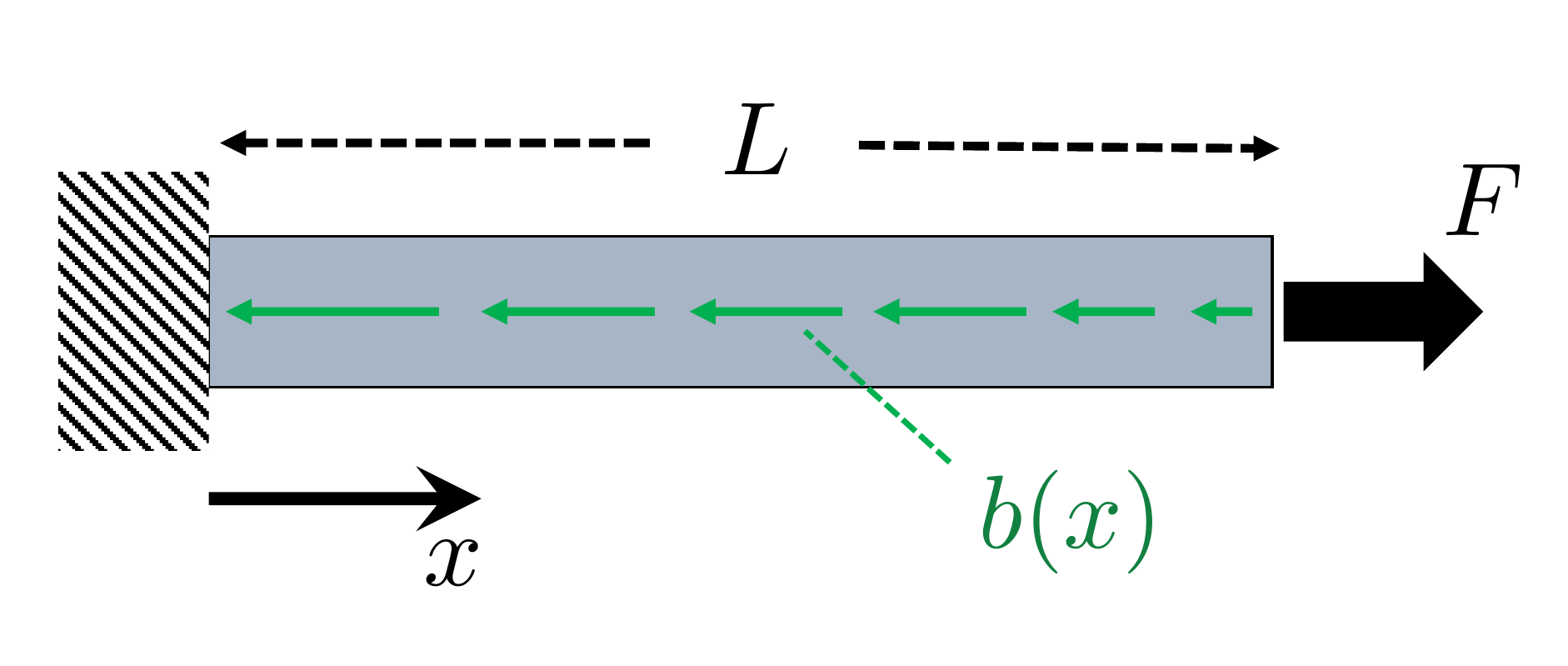}}
  \caption{One-dimensional bar with length $L$, body force $b(x)$, and applied force $F$. We use 50 uniform finite elements with linear basis functions to discretize the domain.
   \label{fig::1d-bar-domain}}
\end{figure}

In this problem, we compare solutions obtained by the proposed method and the original distance minimizing method \citep{kirchdoerfer2016data} for a 1D bar domain that has the following analytical solution:
\begin{equation}
    u(x) = 0.01 x^2; x \in [0, L], 
\end{equation}
where the bar length $L=1$m, shown in \fig \ref{fig::1d-bar-domain}. The applied force at the right end is set to $F = 833.6546$ N, and the left end is fixed from movement. The cross-sectional area of the bar is constant and equal to $1\ \text{mm}^2$. 
\subsubsection{Limited, complete database}\label{subsec::pr1-comp-data}
The material database is synthesized from a known nonlinear model as follows:
\begin{equation}
    \sigma(\epsilon) = \alpha_m \tanh(\alpha_s \epsilon), 
\end{equation}
where $\alpha_m = 1000$ MPa and $\alpha_s=60$ are material parameters. The database is populated by sampling 41 equally spaced data points in the range $\epsilon \in [-0.03, 0.03]$. For comparison purposes, we refer it as the \textit{complete} database.

The original distance minimization method directly uses the database in its local optimization step. However, the method introduced in this paper requires an appropriate mapping $\mathcal{F}: \mathbb{R}^2 \mapsto \mathbb{R}^2$ to perform operations between the ambient data space and the mapped vector space, as explained in Sections \ref{sec::new-local} and \ref{sec::nn-embd}. This mapping is found in an offline manner by a single invertible layer which has three hidden \texttt{elu} layers \citep{clevert2015fast} with five hidden units per layer. Neural network weights and biases are initialized by the uniform Kaiming approach \citep{he2015delving}.

\begin{figure}[h]
 \centering
{\includegraphics[width=0.4\textwidth]{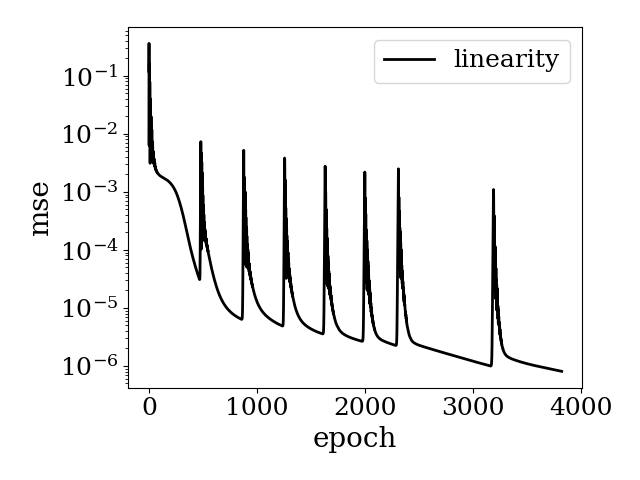}}
  \caption{Mean squared error (mse) of the linearity condition violation during the training epochs for \textit{complete} data shown in \fig \ref{fig::pr1-database}(a). The training is performed in a full-batch manner.
   \label{fig::pr1-loss-trn}}
\end{figure}

The neural network parameters are found by \texttt{ADAM} optimizer \citep{kingma2014adam} with the initial learning rate set at 0.05. We use \texttt{ReduceLROnPlateau} learning scheduler of \texttt{PyTorch} library \cite{paszke2019pytorch} to adjust the learning rate every 50 iterations after the first 2000 iterations. The learning rate reduction factor is set to 0.91 with the minimum learning rate 1e-6. Before training, strain and stress data points are linearly normalized based on their maximum and minimum values to be positive and less than or equal to 1. The training performance is shown in \fig \ref{fig::pr1-loss-trn}.

\begin{figure}[h]
 \centering
 \subfigure[material database]
{\includegraphics[width=0.4\textwidth]{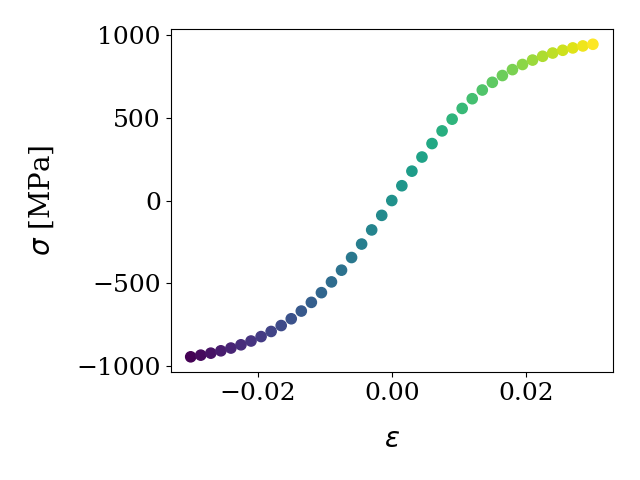}}
\hspace{0.01\textwidth}
 \subfigure[mapped material database]
{\includegraphics[width=0.4\textwidth]{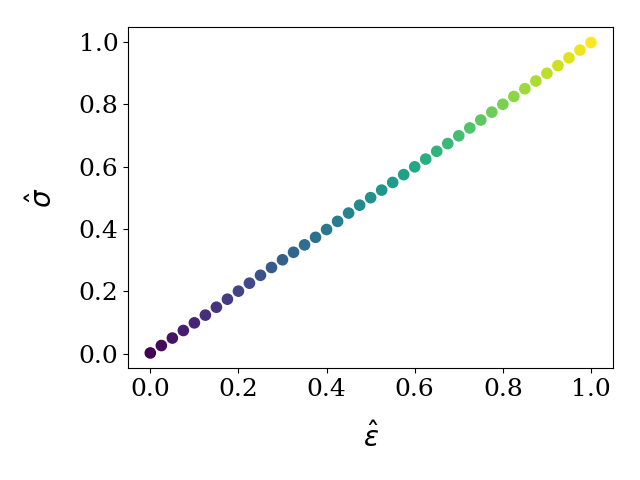}}
\hspace{0.01\textwidth}
  \caption{(a) material database (b) mapped material database into a vector space using the invertible neural network. Colors show the data number.
   \label{fig::pr1-database}}
\end{figure}

\Fig \ref{fig::pr1-database} plots the raw data points and the mapped one which has the vector space properties after training the neural network map function. The results confirm that the proposed neural network architecture is able to find an appropriate bijective function to map forward and backward between the original data space and its vector space counterpart.

\begin{figure}[h]
 \centering
 \subfigure[manifold embedding projection (this work)]
{\includegraphics[width=0.3\textwidth]{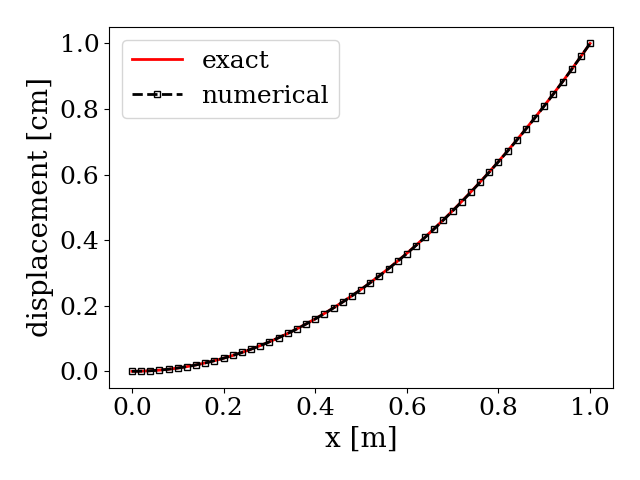}}
\hspace{0.01\textwidth}
 \subfigure[manifold embedding projection (this work)]
{\includegraphics[width=0.3\textwidth]{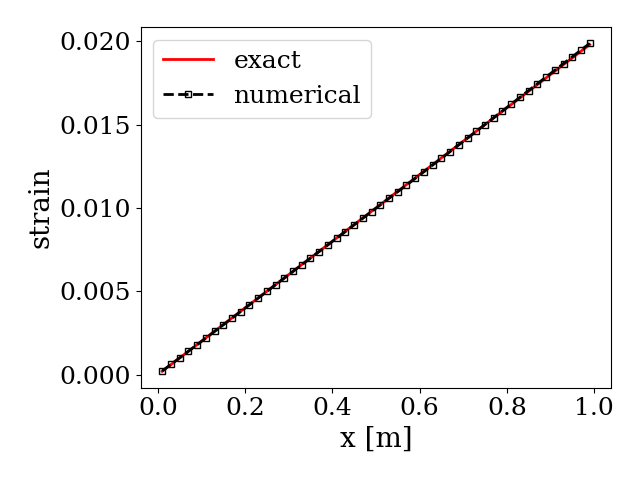}}
\hspace{0.01\textwidth}
 \subfigure[manifold embedding projection (this work)]
{\includegraphics[width=0.3\textwidth]{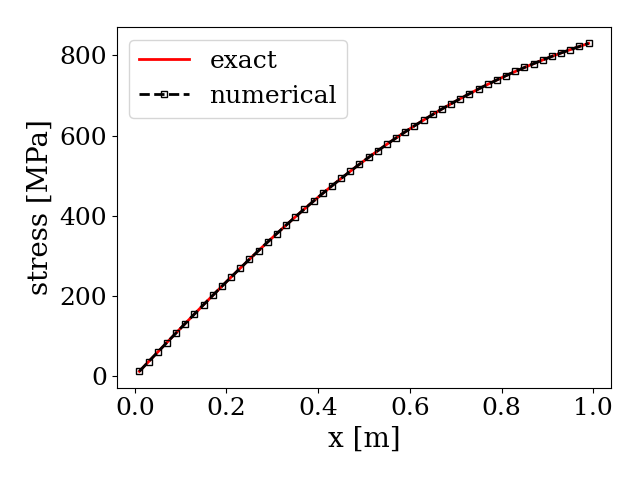}}
\hspace{0.01\textwidth}
 \subfigure[nearest neighbour projection]
{\includegraphics[width=0.3\textwidth]{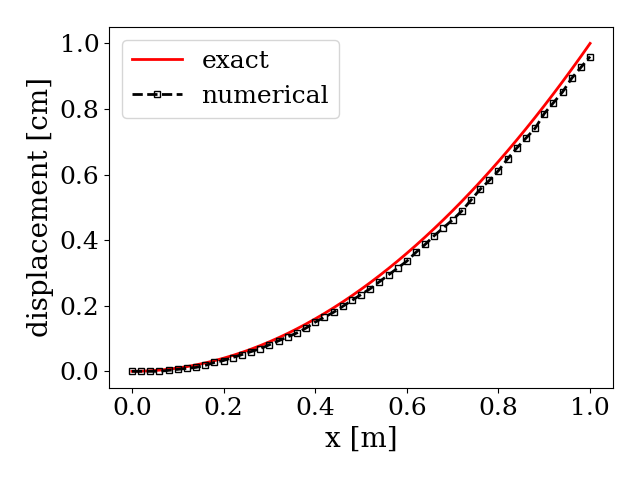}}
\hspace{0.01\textwidth}
 \subfigure[nearest neighbour projection]
{\includegraphics[width=0.3\textwidth]{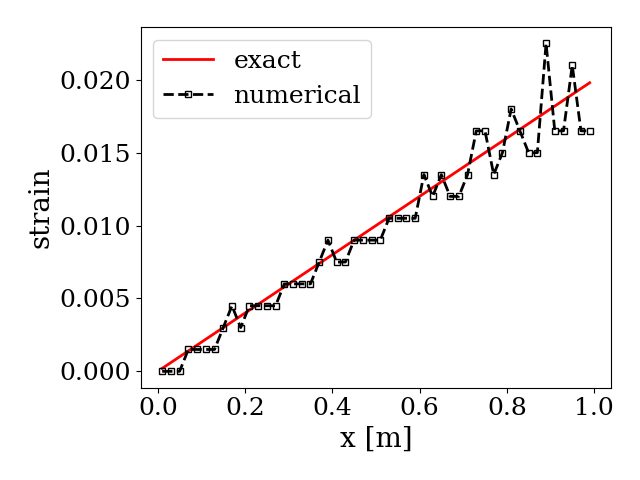}}
\hspace{0.01\textwidth}
 \subfigure[nearest neighbour projection]
{\includegraphics[width=0.3\textwidth]{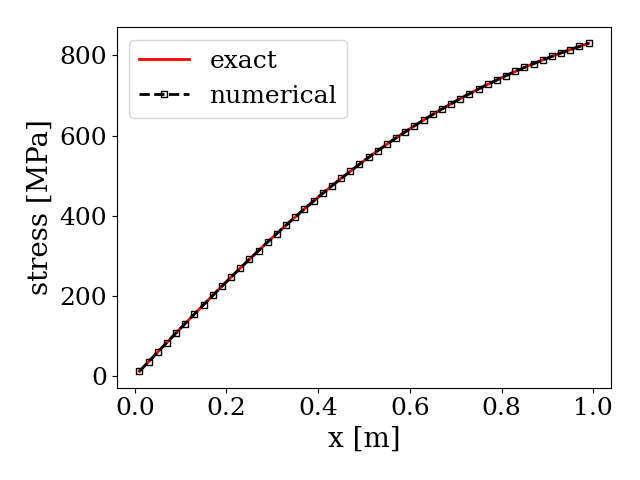}}
\hspace{0.01\textwidth}
  \caption{A comparison between solution fields obtained by the proposed scheme and nearest neighbor projection onto the database. The strain field predictions by the proposed method are more accurate and smooth; see (b) and (e). 
   \label{fig::pr1-sols}}
\end{figure}

We compare the distribution of displacement, strain, and stress along the bar between the proposed scheme and the original distance minimization method in \fig \ref{fig::pr1-sols}. Both schemes have a good agreement with the exact solution. However, the strain field prediction by the original distance minimization method is less accurate and suffers from non-smooth behavior. This observation is not unexpected since the amount of data is not sufficient for the original distance minimization method \citep{kirchdoerfer2016data}. Notice that the original distance minimization method utilizes an energy norm to measure the distance which is equivalent to the l2 norm; see \cite{bahmani2021kd} which introduces an isometric projection between energy norm and l2 norm. However, it is classically known that solid constitutive behavior belongs to a nonlinear smooth manifold that doesn't have the Euclidean vector space structure. This may justify why this method suffers from inaccurate and or non-smooth predictions in the limited data regime. In the big data regime, even an l2 metric can measure the underlying geodesic structure sufficiently well.

\begin{figure}[h]
 \centering
 \subfigure[manifold embedding projection (this work)]
{\includegraphics[width=0.4\textwidth]{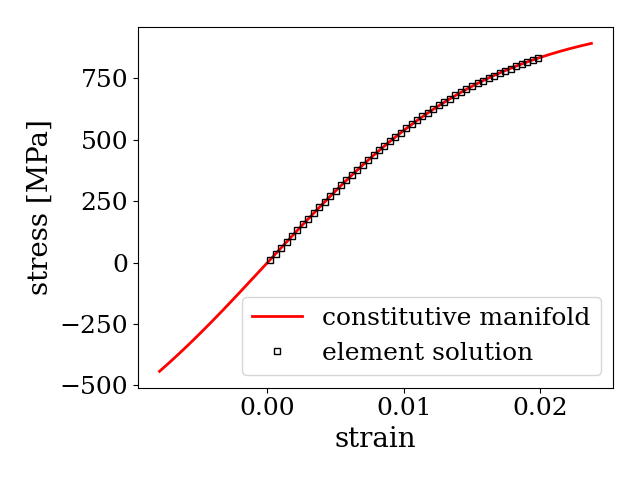}}
\hspace{0.01\textwidth}
 \subfigure[nearest neighbour projection]
{\includegraphics[width=0.4\textwidth]{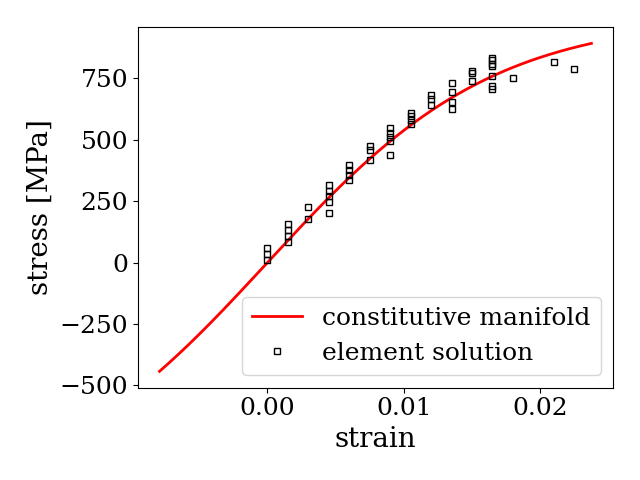}}
\hspace{0.01\textwidth}
  \caption{Strain-stress values obtained via (a) the introduced projection and (b) the nearest neighbor projection. The proposed scheme can identify material states that genuinely belong to the underline hidden material manifold.
   \label{fig::pr1-strs-strn-cov}}
\end{figure}

As a support for the previous claim, we plot the strain-stress state at each bar element for both schemes in \fig \ref{fig::pr1-strs-strn-cov}. The found material states by the proposed scheme accurately resemble the underlying generative function that populates the material database. This observation indicates that the proposed scheme is able to indirectly learn and recover the underlying geometric structure of the material behavior.

\subsubsection{Limited, incomplete database}

\begin{figure}[h]
 \centering
 \subfigure[material database]
{\includegraphics[width=0.4\textwidth]{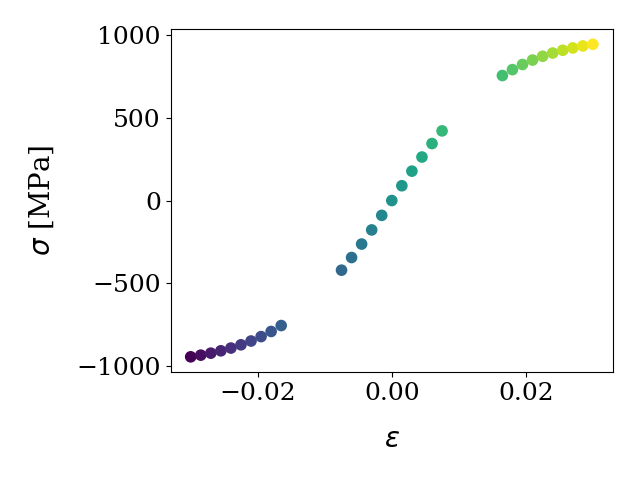}}
\hspace{0.01\textwidth}
 \subfigure[mapped material database]
{\includegraphics[width=0.4\textwidth]{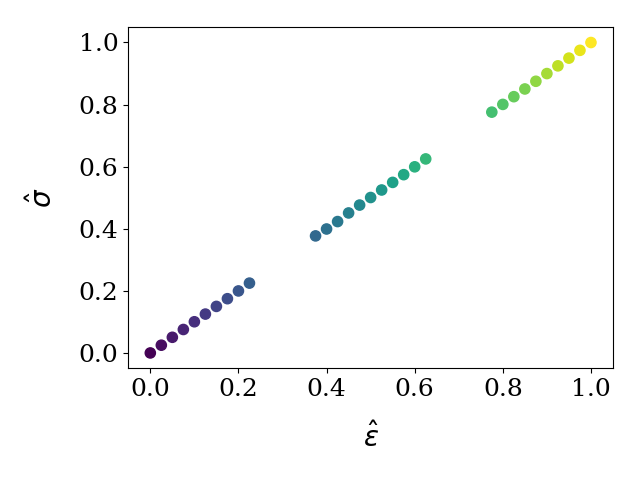}}
\hspace{0.01\textwidth}
  \caption{(a) \textit{incomplete} material database (b) mapped material database into a vector space using the invertible neural network. Colors show the data number.
   \label{fig::pr1-data-incomp}}
\end{figure}

\begin{figure}[h]
 \centering
{\includegraphics[width=0.4\textwidth]{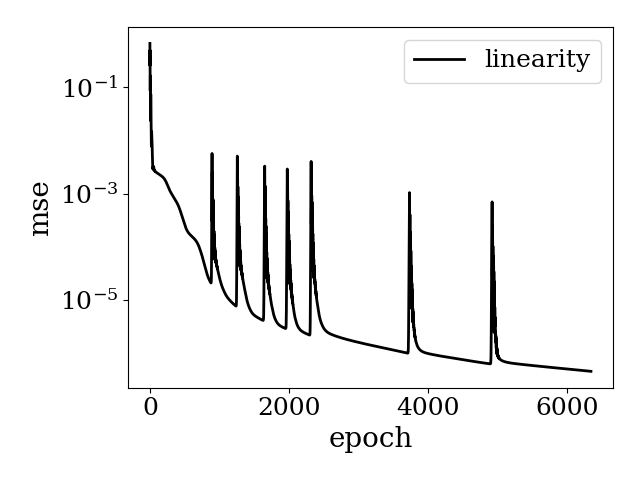}}
  \caption{Mean squared error (mse) of the linearity condition violation during the training epochs for \textit{incomplete} data shown in \fig \ref{fig::pr1-data-incomp}(a). The training is performed in a full-batch manner.
   \label{fig::pr1-loss-trn-incomp}}
\end{figure}

To validate the robustness of the proposed scheme, we eliminate some data points from the nonlinear regions of the previous database and call the new database \textit{incomplete}, shown in \fig \ref{fig::pr1-data-incomp}(a). We retrain the neural network map function for the new database, and the training performance is reported in \fig \ref{fig::pr1-loss-trn-incomp}; the architecture and hyper-parameters are kept the same as the ones used for the \textit{complete} database.

\begin{figure}[h]
 \centering
 \subfigure[manifold embedding projection (this work)]
{\includegraphics[width=0.3\textwidth]{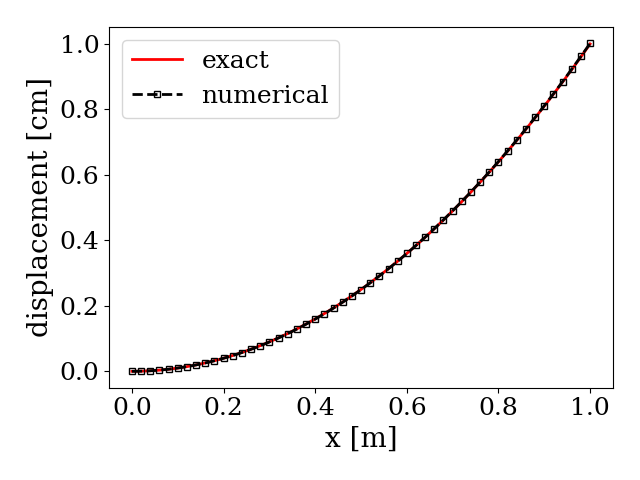}}
\hspace{0.01\textwidth}
 \subfigure[manifold embedding projection (this work)]
{\includegraphics[width=0.3\textwidth]{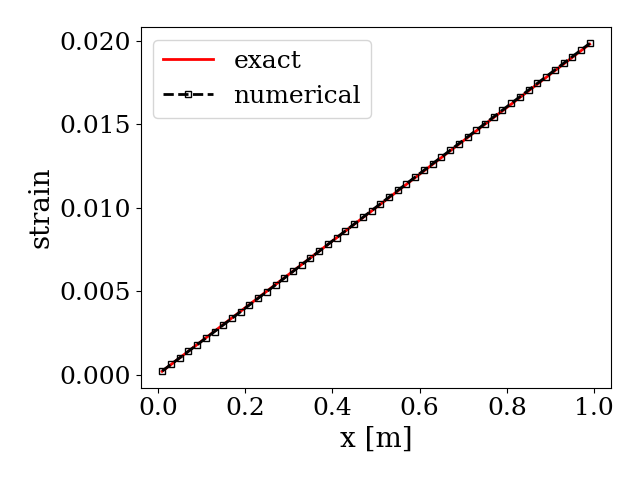}}
\hspace{0.01\textwidth}
 \subfigure[manifold embedding projection (this work)]
{\includegraphics[width=0.3\textwidth]{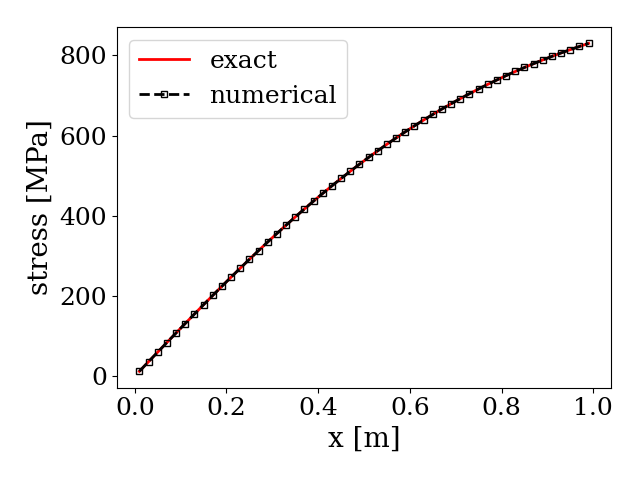}}
\hspace{0.01\textwidth}
 \subfigure[nearest neighbour projection]
{\includegraphics[width=0.3\textwidth]{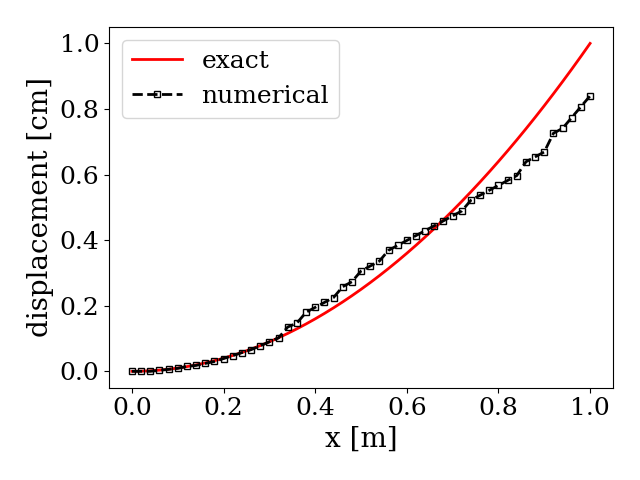}}
\hspace{0.01\textwidth}
 \subfigure[nearest neighbour projection]
{\includegraphics[width=0.3\textwidth]{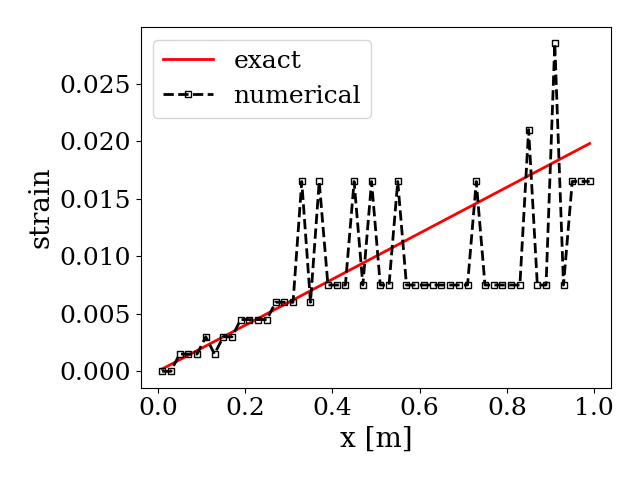}}
\hspace{0.01\textwidth}
 \subfigure[nearest neighbour projection]
{\includegraphics[width=0.3\textwidth]{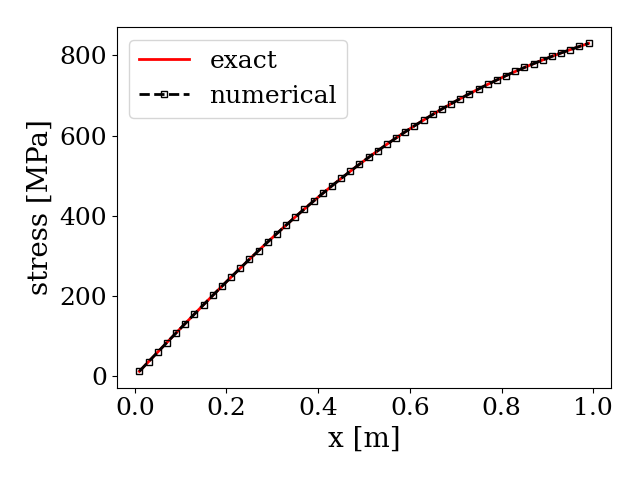}}
\hspace{0.01\textwidth}
  \caption{A comparison between solution fields obtained by the proposed scheme and nearest neighbor projection. Even in the \textit{incomplete} data scenario, the proposed scheme can find more accurate responses compared to the original distance minimization scheme.
   \label{fig::pr1-sol-incompl}}
\end{figure}

The predicted displacement, strain, and stress fields along the bar are shown in \fig \ref{fig::pr1-sol-incompl} for both methods. As the results suggest, the accuracy of the proposed method in this paper is still satisfactory even for such an incomplete and limited database. However, as expected, the accuracy of the original distance minimization is considerably reduced, especially for displacement and strain fields (c.f. \fig \ref{fig::pr1-sols}(d-e)).

\begin{figure}[h]
 \centering
 \subfigure[manifold embedding projection (this work)]
{\includegraphics[width=0.4\textwidth]{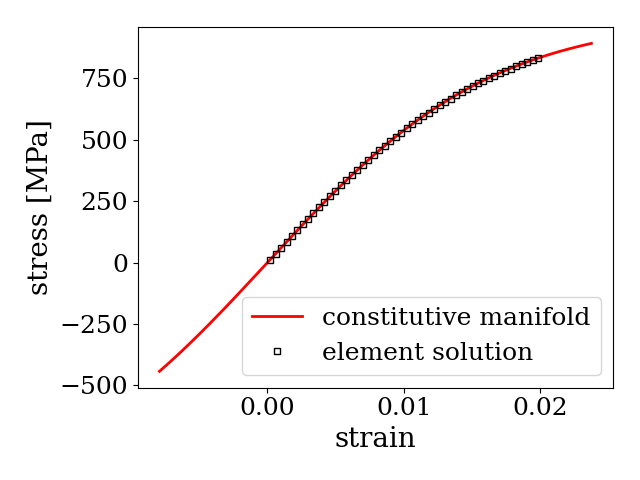}}
\hspace{0.01\textwidth}
 \subfigure[nearest neighbour projection]
{\includegraphics[width=0.4\textwidth]{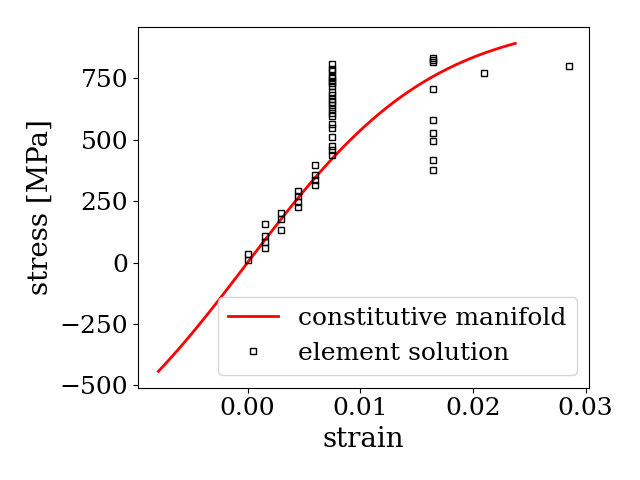}}
\hspace{0.01\textwidth}
  \caption{Strain-stress values obtained at each finite element via (a) the introduced projection and (b) the nearest neighbor projection. The proposed scheme can identify material states that genuinely belong to the underline hidden material manifold, even in the incomplete data-set scenario.
   \label{fig::pr1-strs-strn-cov-incom-data}}
\end{figure}

Similarly, the found strain-stress states in the entire domain are plotted as dot points in \fig \ref{fig::pr1-strs-strn-cov-incom-data} to demonstrate the capabilities of the proposed manifold learning algorithm in recovering the embedded space of the underlying hidden constitutive manifold.
Comparisons of results in  \fig \ref{fig::pr1-strs-strn-cov-incom-data}(a) with \fig \ref{fig::pr1-strs-strn-cov}(a)
reveal that the performance of the proposed manifold method is not significantly affected by the missing data. 
On the other hand, the predicted state values obtained via  the original distance minimization method are shown 
to be very sensitive to the missing data as evidenced in 
\fig \ref{fig::pr1-strs-strn-cov-incom-data}(b) and \fig \ref{fig::pr1-strs-strn-cov}(b). 

This numerical experiment indicates that the knowledge of the geometry of the manifold gained from the 
neural network may improve the robustness of the model-free predictions. Of course, the quality of the projections
may still depend on how the missing data distributes, the smoothness of the data, and whether losing this data may otherwise significantly alter the learned mapping between the constitutive manifold and the embedded hyperplane.

\subsubsection{Comparison with the vanilla autoencoder architecture}
Here, we aim to justify why the proposed architecture may have better performance than a vanilla autoencoder architecture. The details of the autoencoder map function and its optimization statement are described at the beginning of Section \ref{sec::nn-embd}.

Encoder and decoder branches of the autoencoder framework have three hidden layers with three hidden units per layer. For a fair comparison between these two frameworks, this particular network configuration is chosen to be similar to the invertible internal network with almost the same number of unknown parameters. The total number of unknown parameters for the autoencoder and invertible networks are 82 and 76, respectively. The rest of the hyper-parameters are similar to the invertible network reported earlier in subsection \ref{subsec::pr1-comp-data}.

 \begin{figure}[h]
 \centering
{\includegraphics[width=0.4\textwidth]{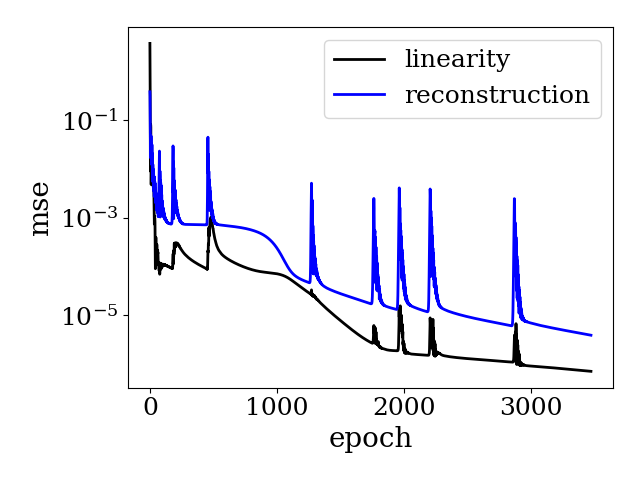}}
  \caption{Mean squared error (mse) of the reconstruction and linearity constraint errors during the training epochs for \textit{complete} data shown in \fig \ref{fig::pr1-data-incomp}(a) using the vanilla autoencoder framework. The training is performed in a full-batch manner. The number of training iterations is similar to the invertible network training shown in \fig \ref{fig::pr1-loss-trn}.
   \label{fig::pr1-loss-trn-encdec}}
\end{figure}

\begin{figure}[h]
 \centering
 \subfigure[introduced invertible architecture]
{\includegraphics[width=0.4\textwidth]{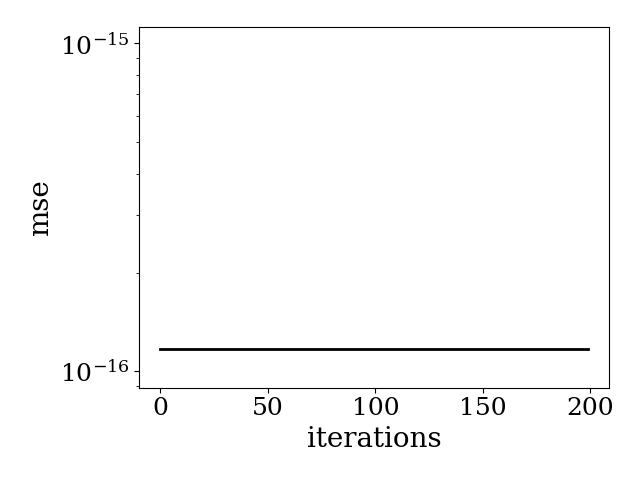}}
\hspace{0.01\textwidth}
 \subfigure[vanilla autoencoder architecture]
{\includegraphics[width=0.4\textwidth]{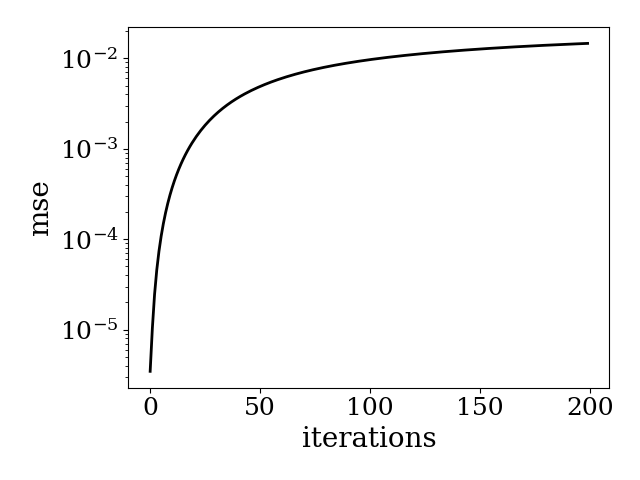}}
\hspace{0.01\textwidth}
  \caption{Comparing the stability of two architectures after 200 forward and backward mappings for 100 randomly generated data points not used in the network training. The reconstruction error of the tailored architecture (a) is almost negligible and remains stable during continuously forward-backward mappings.
   \label{fig::pr1-network-stability}}
\end{figure}

 \Fig \ref{fig::pr1-loss-trn-encdec} shows the autoencoder training performance. With the same number of gradient descent iterations, the obtained error of the linearity constraint is almost similar to the invertible architecture (c.f. \fig \ref{fig::pr1-loss-trn}). However, the autoencoder network's reconstruction error (backward map) is significant relative to the machine precision error of the invertible network, although it is small in the absolute sense. To make this point clearer, we study the reconstruction stability and accuracy of the trained networks during 200 consecutive forward and backward mappings for a batch of 100 data points generated randomly from the constitutive manifold and not seen in the training. The results are plotted in \fig \ref{fig::pr1-network-stability}. Not surprisingly, the accuracy of the invertible network remains constant around the machine precision, while the autoencoder accuracy rapidly decreases up until a saturation level.

\subsection{Nonlinear truss system}\label{subsec::truss}

\begin{figure}[h]
 \centering
{\includegraphics[width=0.5\textwidth]{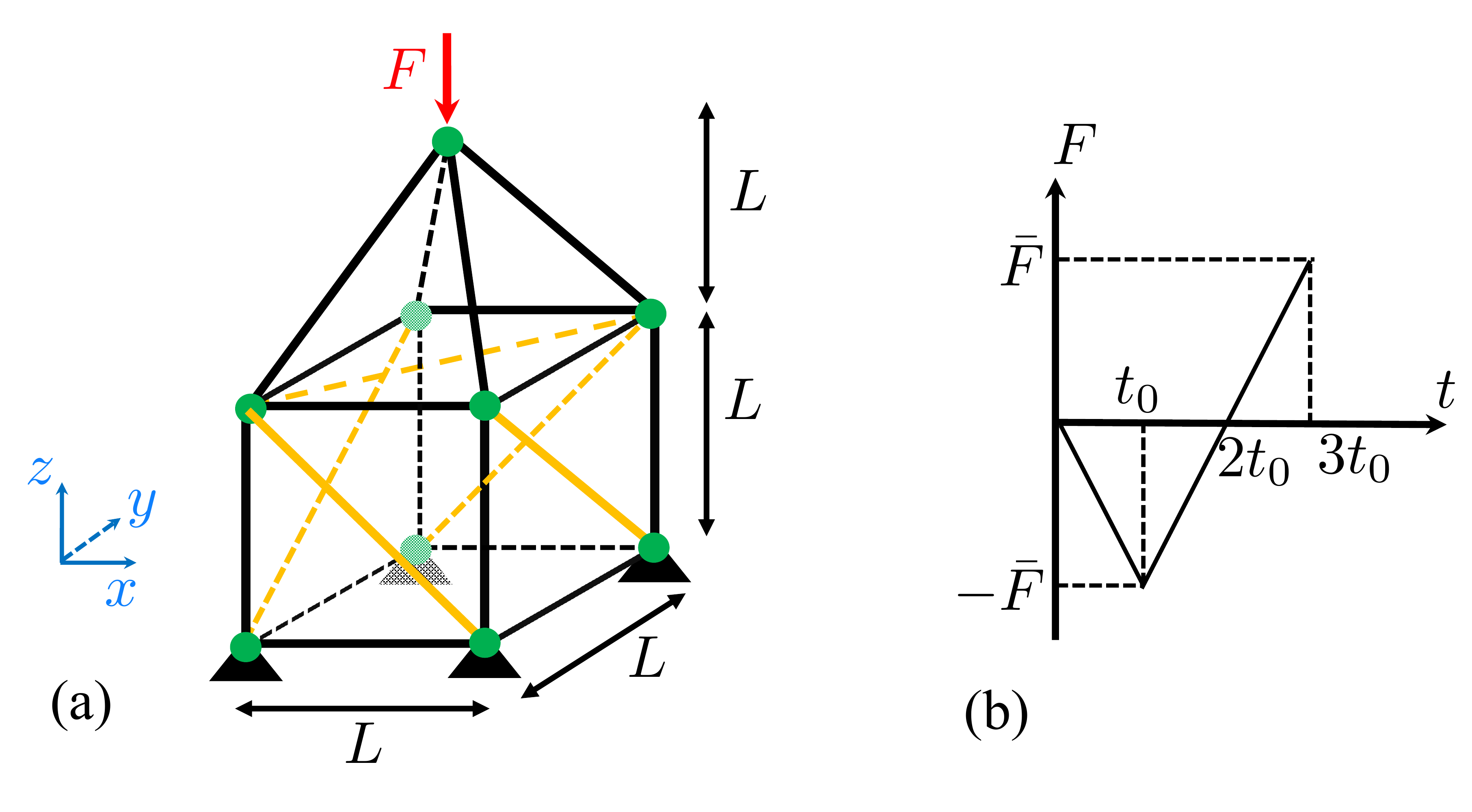}}
  \caption{(a) Truss system under (b) quasi-static cyclic loading $F$. Horizontal axis in (b) indicates the step number. We set $\bar{F}=3$ kN, $t_0=20$, and the entire loading-unloading takes 60 steps.
   \label{fig::3d-truss-domain}}
\end{figure}

In this problem, we compare the efficiency, robustness, and accuracy of the proposed method against the nearest neighbor projection for a 3D truss structure that undergoes quasi-static, elastic loading-unloading, as shown in \fig \ref{fig::3d-truss-domain}. The structure is loading at the top node by a linearly increasing compressive force to reach the maximum force $\bar{F}$, then the compressive force is linearly unloaded up to a maximum tensile force $\bar{F}$ as shown in \fig \ref{fig::3d-truss-domain}(b). The continuum elasticity equations can be straightforwardly simplified for 3D truss elements, hence the details of the derivation are omitted for the sake of brevity; interested readers can follow reference books such as \citet{jacob2007first,belytschko2014nonlinear}.

The material database is similar to the complete data used in the previous problem (see \fig \ref{fig::pr1-database}), and we use the same neural network map function as the previous problem. The global optimization parameters and local ones (in the case of the original distance minimization method) are the same as the previous problem, i.e., $C = S^{-1} = 42694.67$MPa.

\begin{figure}[h]
 \centering
 \subfigure[manifold embedding projection (this work)]
{\includegraphics[width=0.4\textwidth]{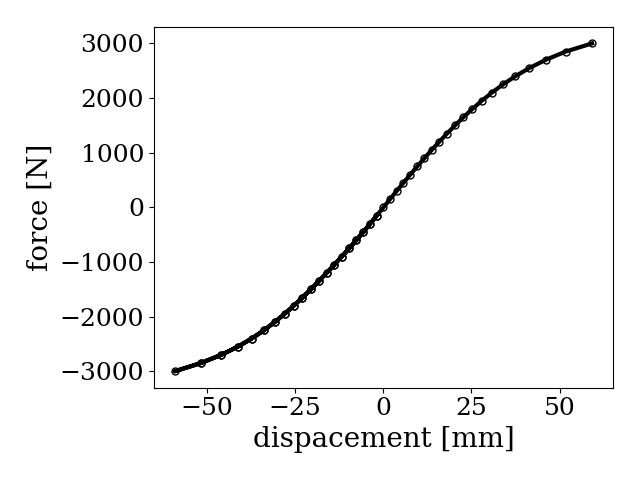}}
\hspace{0.01\textwidth}
 \subfigure[nearest neighbour projection]
{\includegraphics[width=0.4\textwidth]{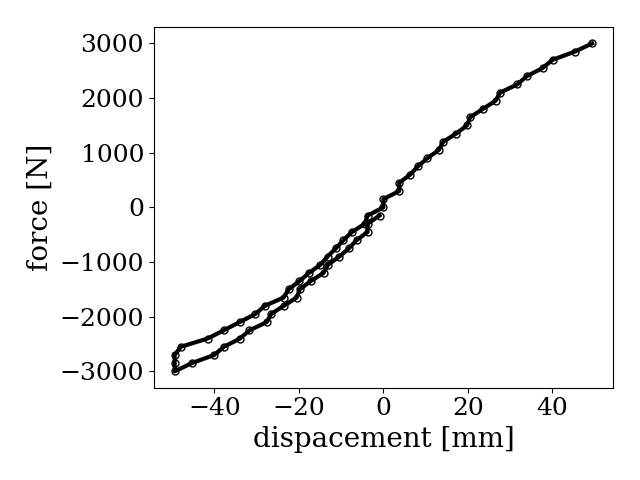}}
\hspace{0.01\textwidth}
  \caption{force-displacement relation for the top node during the loading history via (a) the introduced projection and (b) the nearest neighbor projection. The introduced approach does not suffer from a possible spurious history dependence for an elastic material, in the limited data regime.
   \label{fig::pr2-force-disp}}
\end{figure}

\Fig \ref{fig::pr2-force-disp} depicts the force-displacement plots obtained by the proposed data-driven scheme and the original distance minimization method. Interestingly, the original distance minimization method shows spurious hysteresis behavior in the unloading stage which could be wrongly considered as an irreversible mechanism symptom by a not experienced practitioner, while the proposed scheme in this paper accurately predicts the smooth elastic behavior.

\begin{figure}[h]
 \centering
 \subfigure[introduced projection]
{\includegraphics[width=0.4\textwidth]{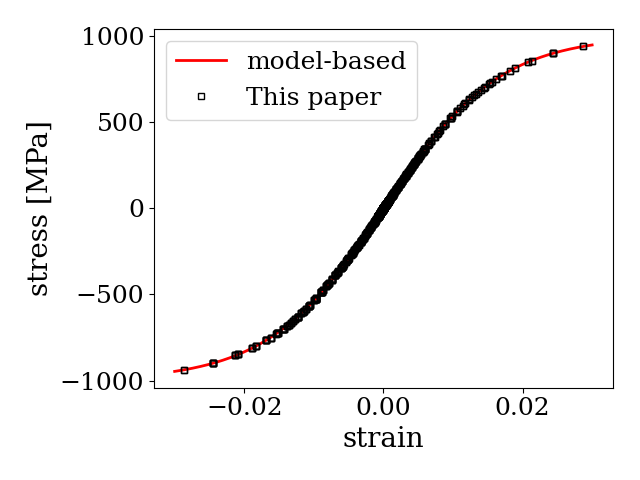}}
\hspace{0.01\textwidth}
 \subfigure[nearest neighbour projection]
{\includegraphics[width=0.4\textwidth]{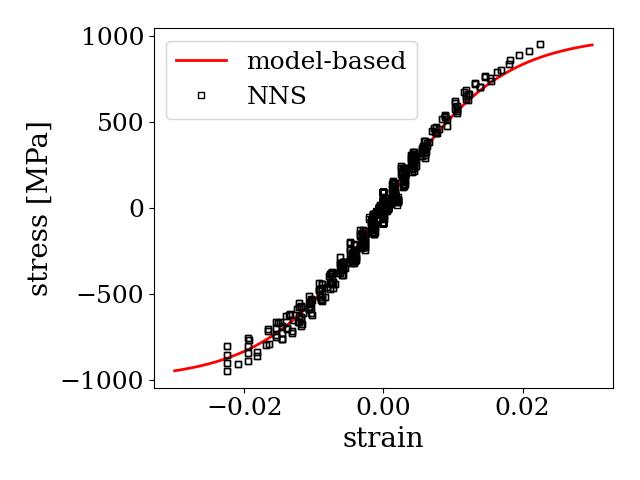}}
\hspace{0.01\textwidth}
  \caption{Strain-stress values obtained at each finite element during the loading history via (a) the introduced projection and (b) the nearest neighbor projection. The founded material states by the introduced scheme (a) follow the underlying, hidden material manifold, especially in the nonlinear regions.
   \label{fig::pr2-strs-strn-cov}}
\end{figure}

The strain-stress pairs in each bar element in the truss system during the entire loading history are plotted in \fig \ref{fig::pr2-strs-strn-cov} to investigate the offset between the estimated material states (dot points) and the underlying material constitutive manifold (red line). As these plots suggest, the proposed scheme finds the material states that are truly on the underlying material manifold. However, the original distance minimization method cannot precisely recover the nonlinear behavior. Notice that this performance issue of the original distance minimization is not unexpected as it is known as a data demanding method due to its minimal assumption about the constitutive model \citep{bahmani2021kd}. In other words, it can converge to the true behavior in the big data regimes \citep{kirchdoerfer2016data,conti2018data}.

\begin{figure}[h]
 \centering
  \subfigure[model-based]
{\includegraphics[width=0.3\textwidth]{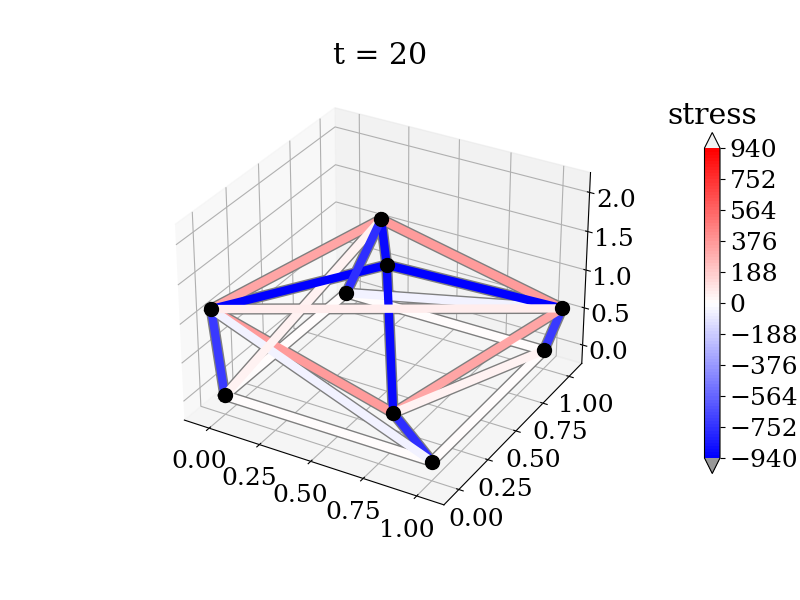}}
\hspace{0.01\textwidth}
 \subfigure[manifold embedding projection (this work)]
{\includegraphics[width=0.3\textwidth]{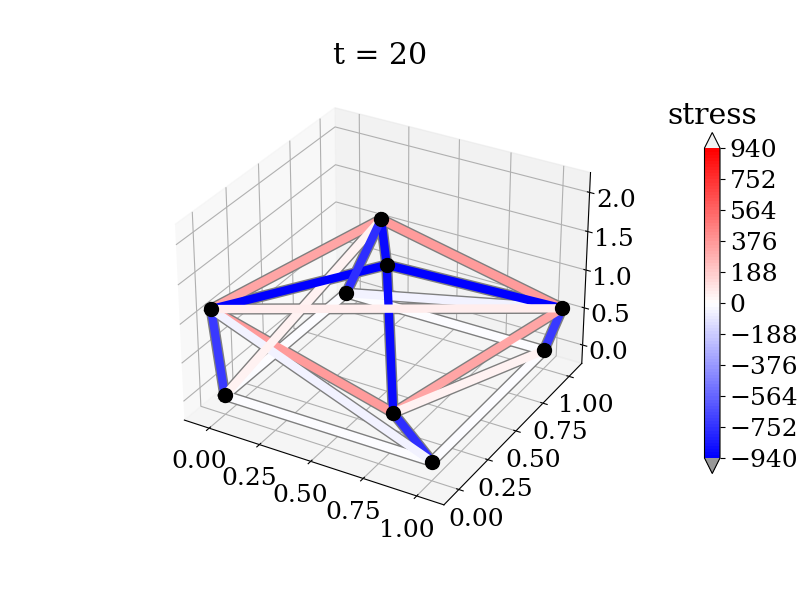}}
\hspace{0.01\textwidth}
 \subfigure[nearest neighbour projection]
{\includegraphics[width=0.3\textwidth]{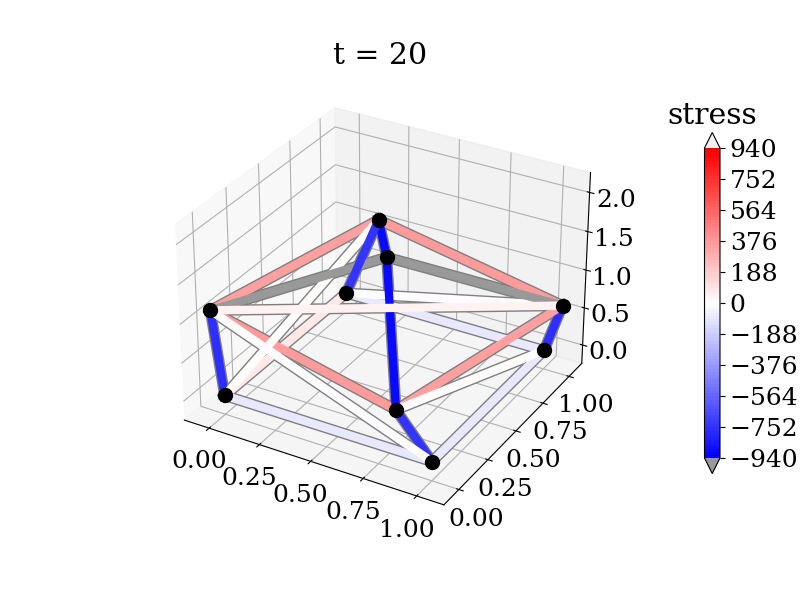}}
\hspace{0.01\textwidth}
 \subfigure[model-based]
{\includegraphics[width=0.3\textwidth]{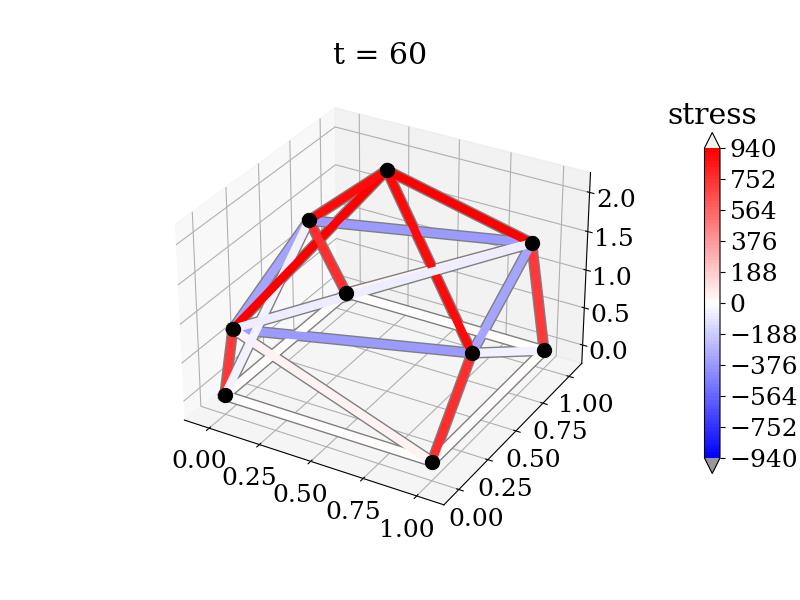}}
\hspace{0.01\textwidth}
 \subfigure[manifold embedding projection (this work)]
{\includegraphics[width=0.3\textwidth]{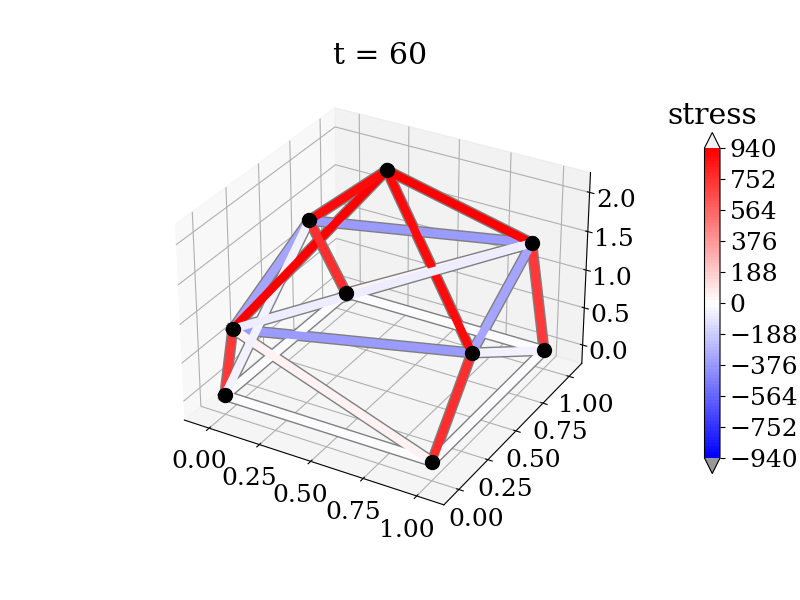}}
\hspace{0.01\textwidth}
 \subfigure[nearest neighbour projection]
{\includegraphics[width=0.3\textwidth]{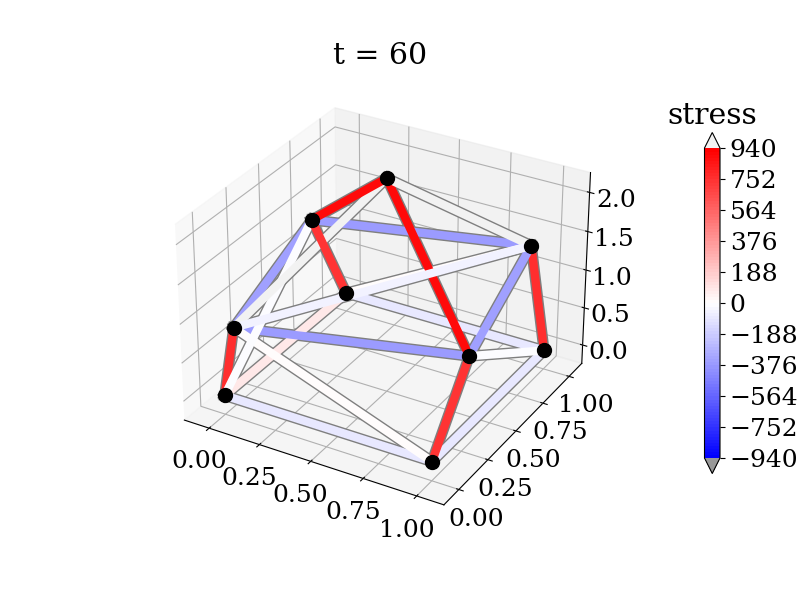}}
\hspace{0.01\textwidth}
  \caption{A comparison between solution fields obtained by the proposed scheme and nearest neighbor projection onto the database. In (c) and (f), the brute-force distance minimization scheme provides wrong stress states for the top left and right bars.
   \label{fig::pr2-strs-def}}
\end{figure}

The stress values in each bar element for the maximum compressive and tensile applied forces are shown in \fig \ref{fig::pr2-strs-def} which are obtained by three solvers: classical model-based method, the introduced projection method herein, and the original distance minimization method. There is a good agreement between the model-based method and the method introduced in this paper, while the original distance minimization method predicts inaccurate stress values in some elements; e.g., see left and right elements of the top Pyramid.

\subsection{Nonlinear conduction}\label{subsec::heat}

\begin{figure}[h]
 \centering
{\includegraphics[width=0.3\textwidth]{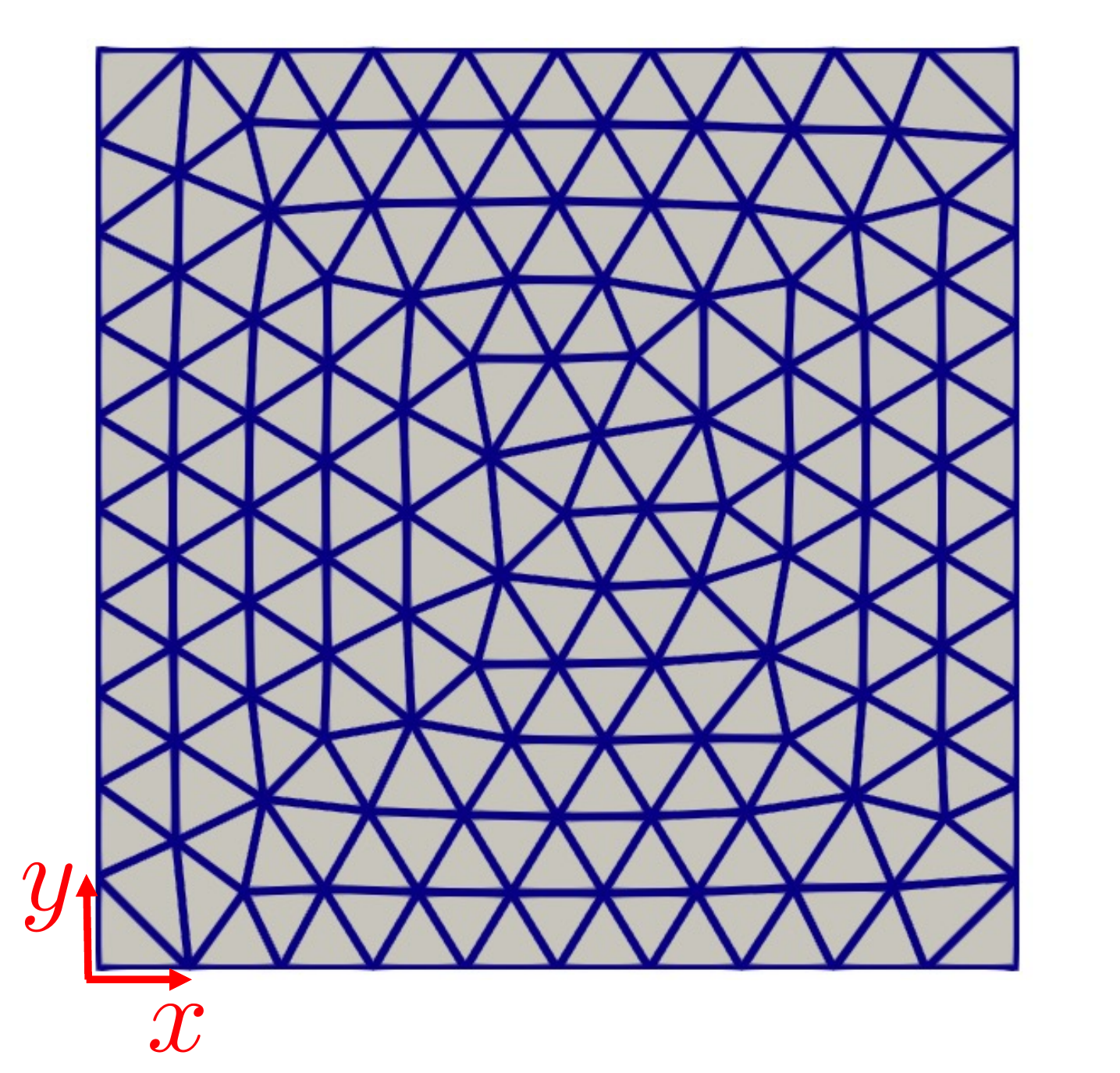}}
  \caption{The square domain of length 1m for the heat conduction problem with pure Dirichlet boundary and non-zero heat source. The domain is discretized by 226 triangular finite elements.
   \label{fig::pr3-domain}}
\end{figure}

Here, we showcase the application of the proposed data-driven method for nonlinear heat transfer problems. To compare the robustness of the method in the limited data scenario, we solve the problem posed in \citep{nguyen2020variational} which admits the following analytical solution for a square domain shown in \fig \ref{fig::pr3-domain},
\begin{equation}
    T(x, y) = \frac{1}{2} \sin{(2\pi x)} y (1-y).
\end{equation}
Similar to \citep{nguyen2020variational}, the database is synthesized by sampling 400 data points from the following constitutive behavior,
\begin{equation}
    \vec{q} = \tanh{(\grad T)},
\end{equation}
where $\vec{q}$ is the heat flux vector. For the sake of brevity, the details of derivation for Poisson's equation are omitted, and interested readers can straightforwardly extend the method introduced here to Poisson's equation, see also \citep{nguyen2020variational, bahmani2021kd}.

To populate the database, a regular grid is used for temperature gradient components in the ranges $-1 \le \partial T / \partial x \le 1$ and $-1 \le \partial T / \partial y \le 1$. The $\mat{C}$ parameter for global optimization is set to $0.42 \tensor{I}$ where $\tensor{I}$ is the second-order identity tensor, and $\mat{S}$ tensor is set to its inverse as suggested in \citep{nguyen2020variational, bahmani2021kd}. The original distance minimization scheme requires similar parameters for the local optimization which is set equal to the global optimization parameters.

For the proposed scheme, we should first find an appropriate map function $\mathcal{F}: \mathbb{R}^4 \mapsto \mathbb{R}^4$ to perform operations between the original data space and the mapped space with the vector space properties. This step is done offline by training an invertible neural network map function. For this database, we design an invertible neural network consisting of four \texttt{elu} layers, each has ten hidden units. The uniform Kaiming approach initializes the network weights and biases.

\begin{figure}[h]
 \centering
{\includegraphics[width=0.5\textwidth]{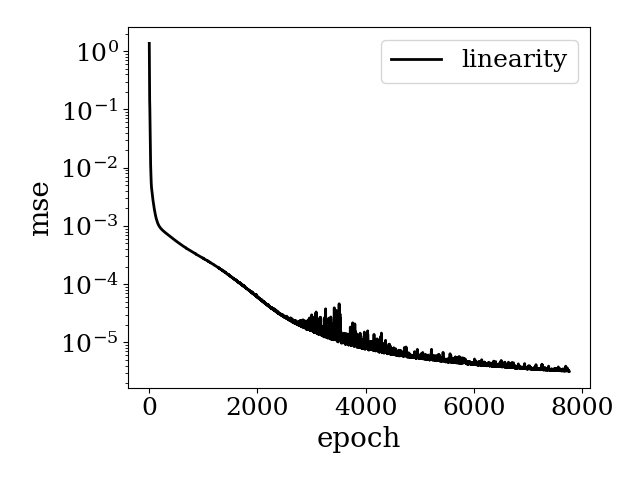}}
  \caption{Mean squared error (mse) of the linearity condition violation during the training epochs for the 2D nonlinear heat transfer data. Random mini-batches of the size 200 are used during the training.
   \label{fig::pr3-loss-trn}}
\end{figure}

\texttt{ADAM} optimizer iterates to find close to optimal parameters for the mapping function with an initial learning rate=0.002. We use \texttt{ReduceLROnPlateau} learning scheduler to adjust the learning rate every 50 iterations after the first 500 iterations. The learning rate reduction factor is set to 0.91 with the minimum learning rate 1e-6. Before training, temperature gradient and heat flux data points are linearly normalized based on their maximum and minimum values to be positive and less than or equal to 1. The training performance is shown in \fig \ref{fig::pr3-loss-trn}.

\begin{figure}[h]
 \centering
 \subfigure[manifold embedding projection (this work)]
{\includegraphics[width=0.3\textwidth]{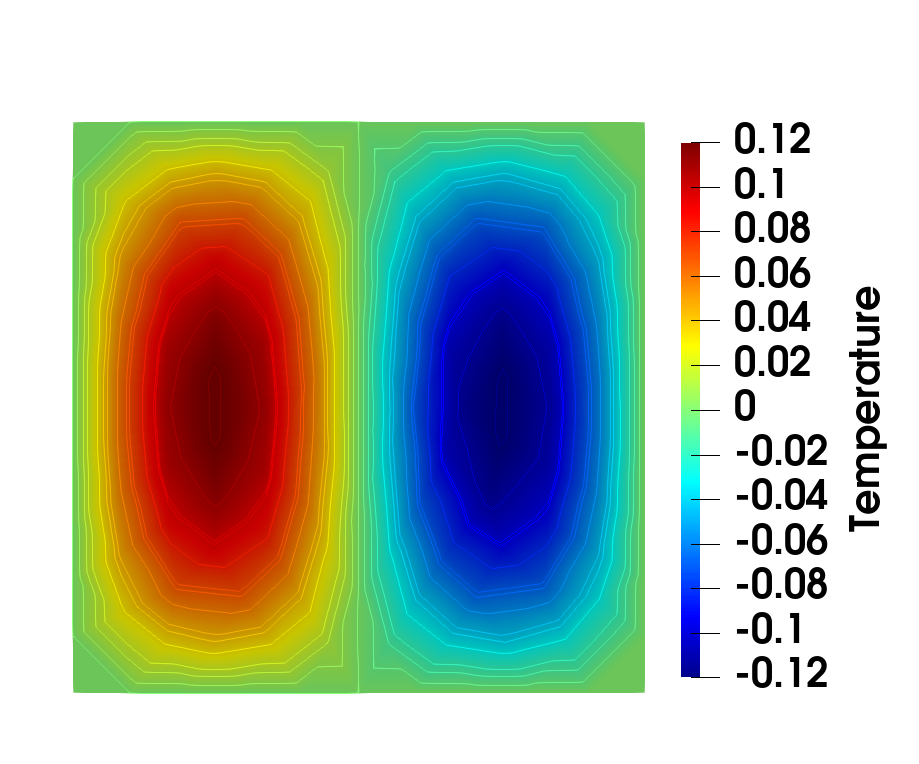}}
\hspace{0.01\textwidth}
 \subfigure[manifold embedding projection (this work)]
{\includegraphics[width=0.3\textwidth]{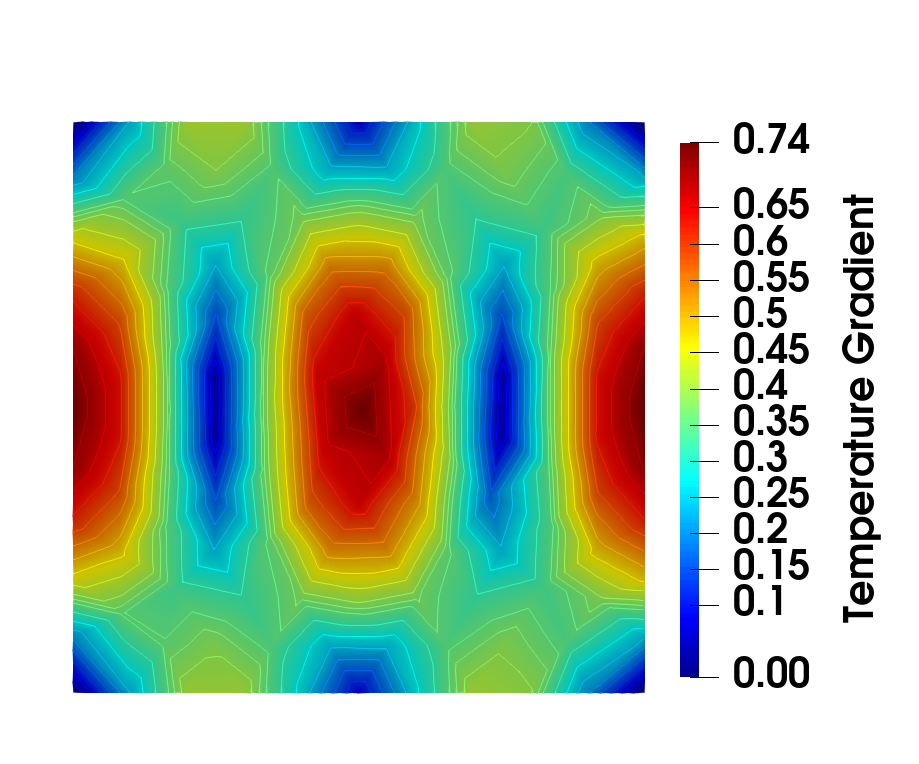}}
\hspace{0.01\textwidth}
 \subfigure[manifold embedding projection (this work)]
{\includegraphics[width=0.3\textwidth]{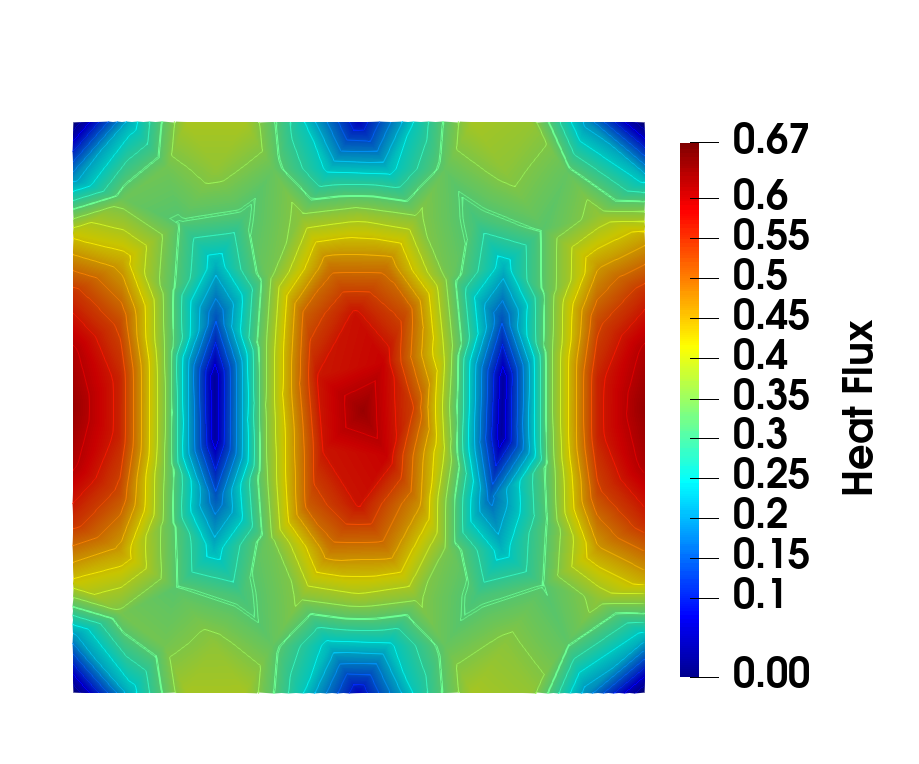}}
\hspace{0.01\textwidth}
 \subfigure[nearest neighbour projection]
{\includegraphics[width=0.3\textwidth]{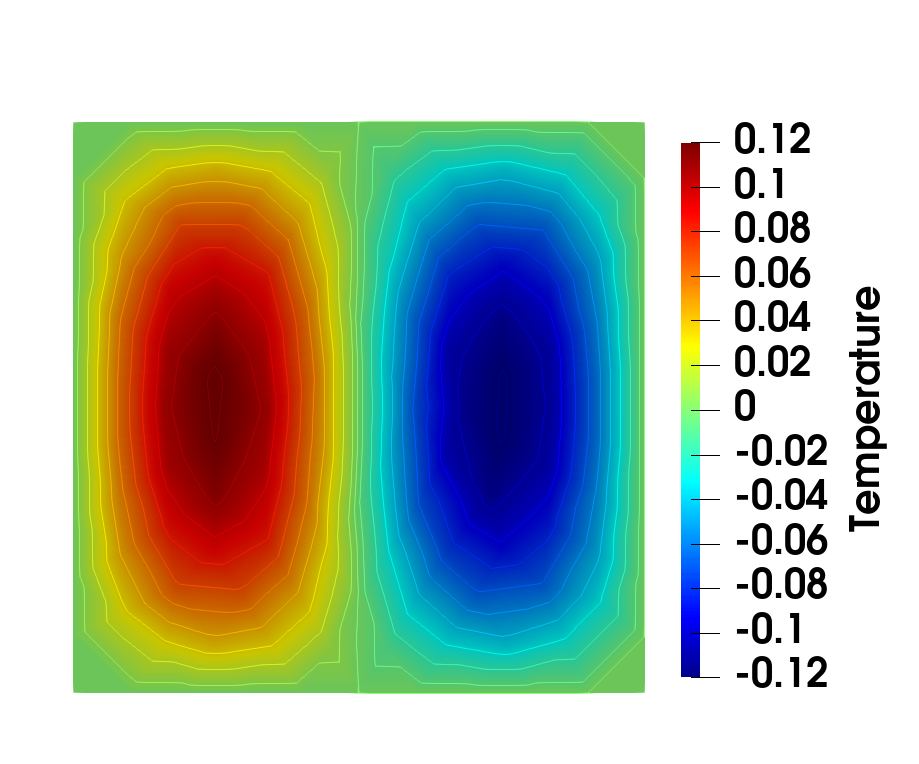}}
\hspace{0.01\textwidth}
 \subfigure[nearest neighbour projection]
{\includegraphics[width=0.3\textwidth]{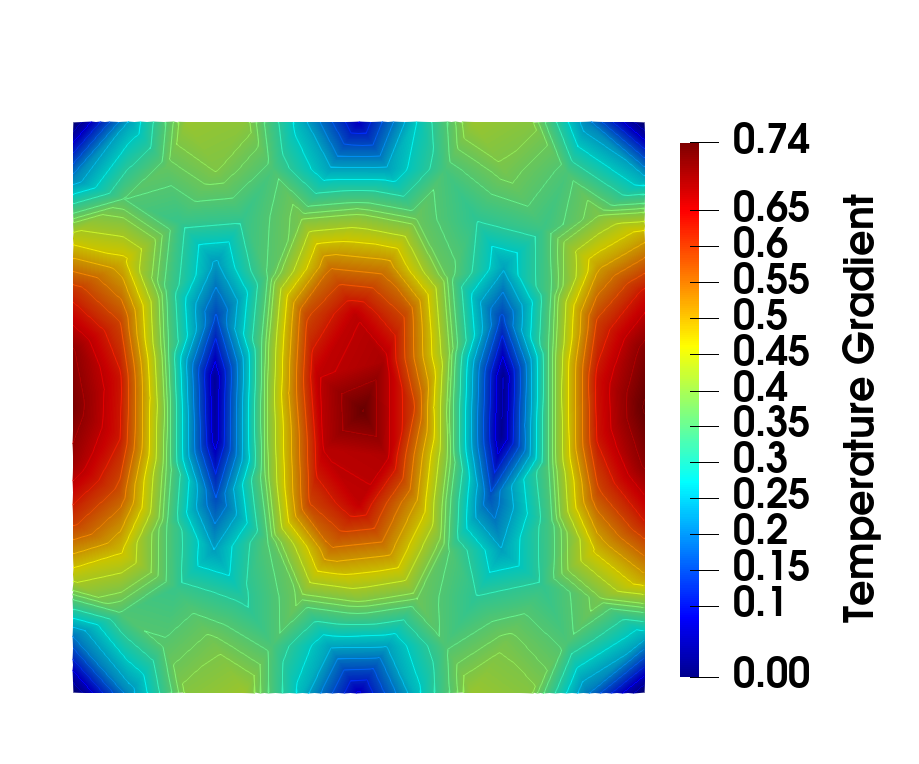}}
\hspace{0.01\textwidth}
 \subfigure[nearest neighbour projection]
{\includegraphics[width=0.3\textwidth]{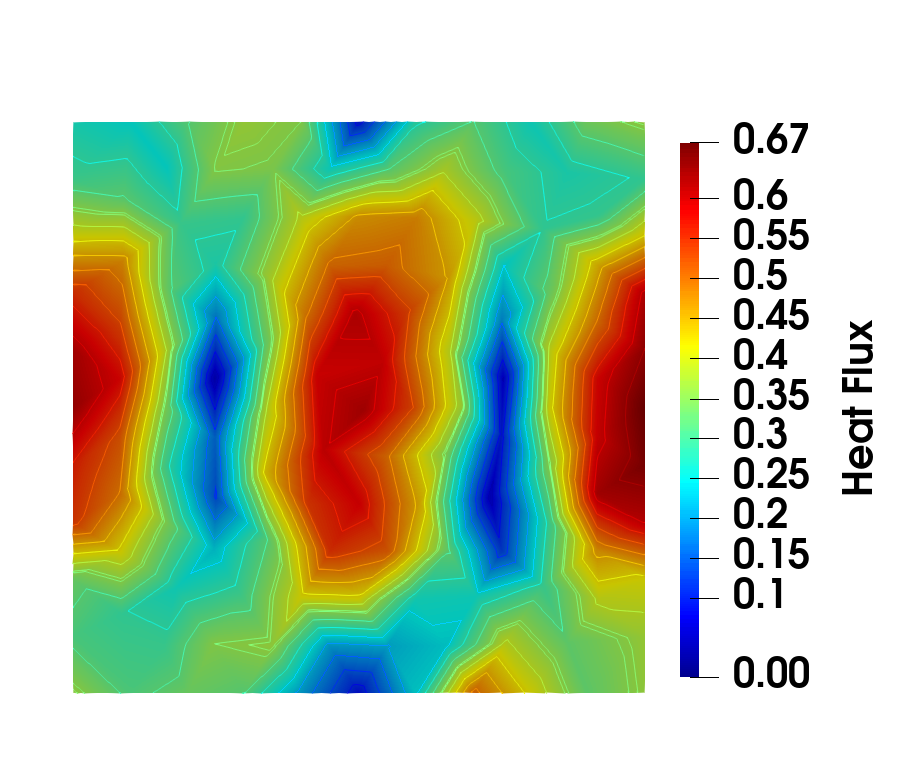}}
\hspace{0.01\textwidth}
\subfigure[model-based]
{\includegraphics[width=0.3\textwidth]{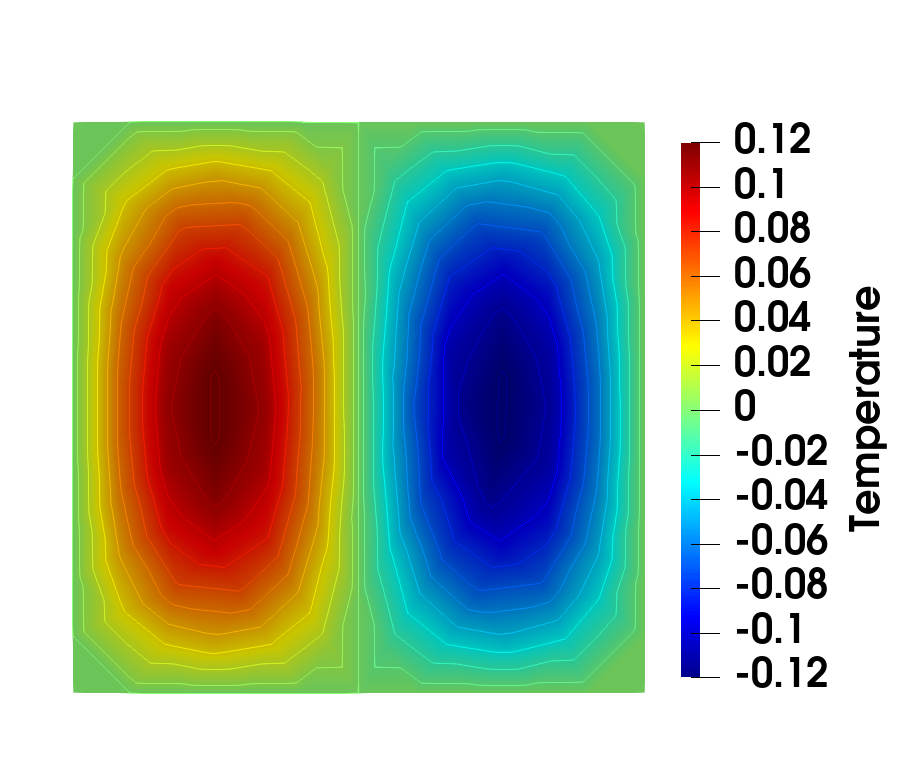}}
\hspace{0.01\textwidth}
 \subfigure[model-based]
{\includegraphics[width=0.3\textwidth]{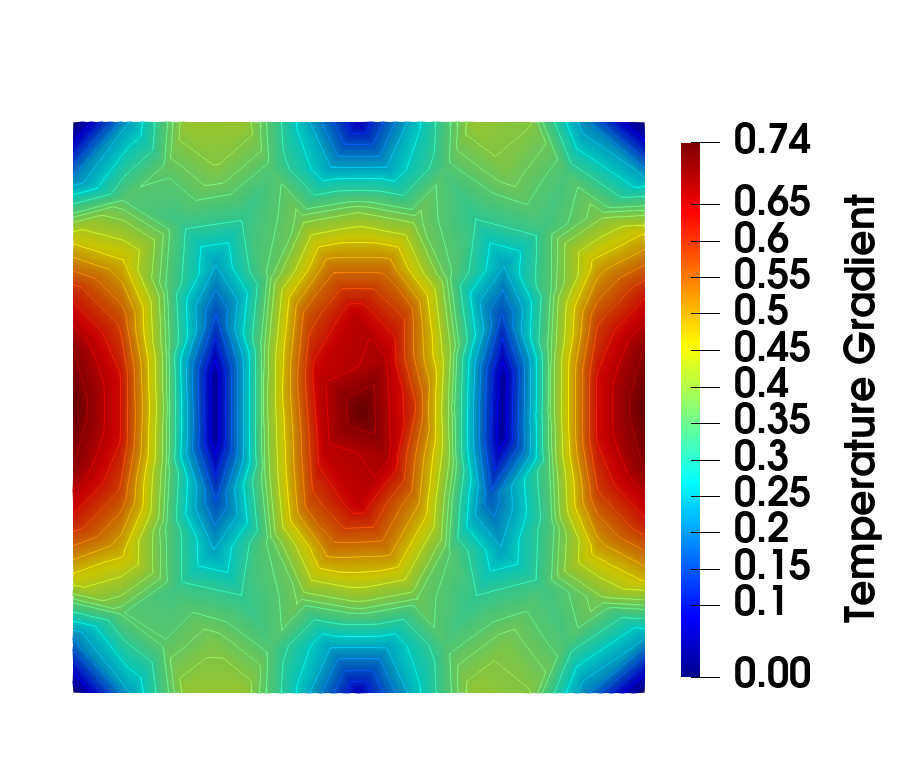}}
\hspace{0.01\textwidth}
 \subfigure[model-based]
{\includegraphics[width=0.3\textwidth]{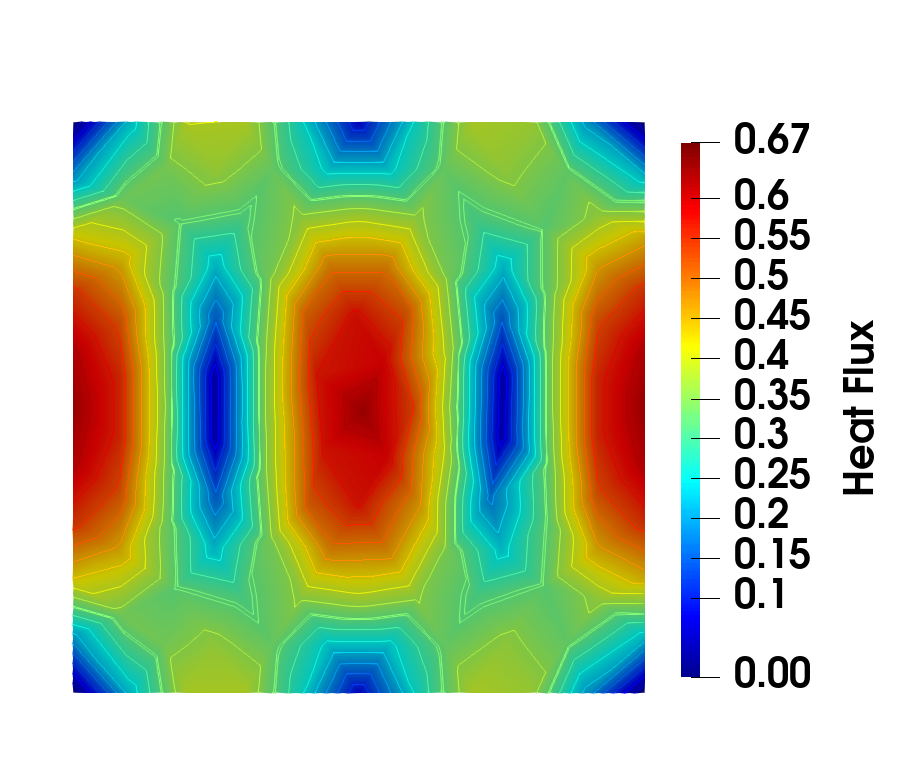}}
\hspace{0.01\textwidth}
  \caption{A comparison between solution fields (temperature, temperature gradient magnitude, and heat flux magnitude) obtained by (a-c) the proposed projection, (d-f) nearest neighbor projection, and (g-i) classical model-based nonlinear finite element solver. The introduced scheme improves heat flux prediction.
   \label{fig::pr3-fields}}
\end{figure}

As shown in \fig \ref{fig::pr3-fields}, there is a good agreement between the proposed scheme and the ground-truth solution (model-based solver). The heat flux predictions by the original distance minimization method are less accurate than the proposed method in the limited data regime. Notice that this superior behavior of the original distance minimization scheme is due to the amount of data used in this study; if the database is sufficiently large, its performance converges to the model-based scheme as well \citep{kirchdoerfer2016data, leygue2018data, nguyen2020variational, bahmani2021kd}.

\subsection{Nonlinear elasticity}\label{subsec::2D-hole}
\begin{figure}[h]
 \centering
{\includegraphics[width=0.4\textwidth]{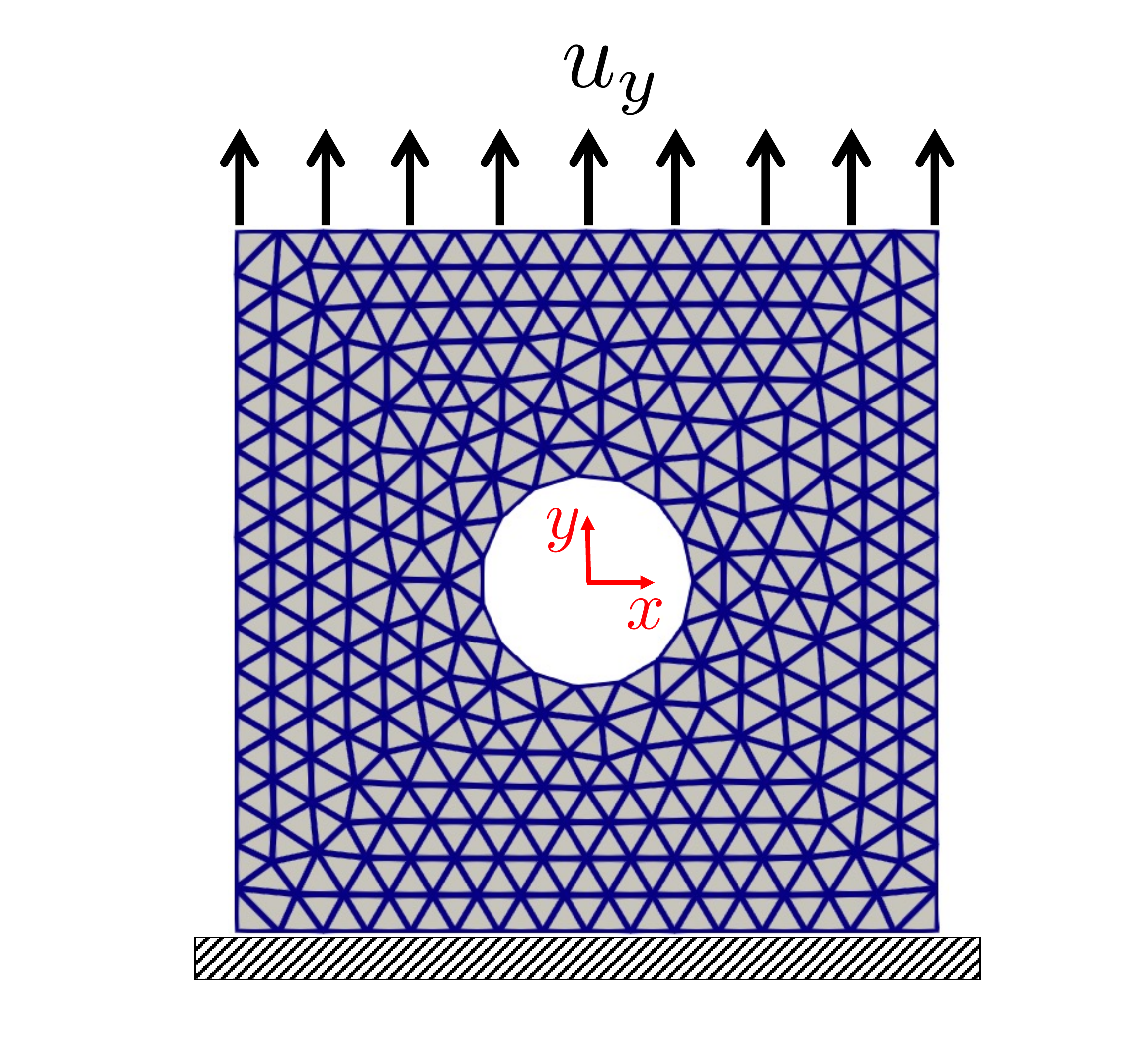}}
  \caption{Plane strain square with length 1m and a circular imperfection at its center with radius 0.15m. The domain is discretized by 545 triangular finite elements.
   \label{fig::pr4-domain}}
\end{figure}

Here, we show the effectiveness of the proposed method compared with the original distance minimization method in dealing with 2D nonlinear elasticity problems. The problem domain and boundary conditions are depicted in \fig \ref{fig::pr4-domain} where the bottom is fixed from horizontal and vertical movements and a uniform vertical displacement $u_y = 0.1$m is applied over the top edge.

We synthesize a database consisting of 1000 data points sampled from the following strain energy \citep{nguyen2020variational,bahmani2021kd}:
\begin{equation}
    \psi(\tensor{\epsilon}) = \frac{1}{2} (1 + 2\epsilon_{kk} - 2 \log{(1 + \epsilon_{kk})}) + \frac{1}{2} (\log{(1+\epsilon_{kk})})^2 + \epsilon_{ki} \epsilon_{ik}.
\end{equation}
Plane strain condition is assumed and a regular grid is used to sample the strain components in the ranges $-0.335 \le \epsilon_{11} \le 0.0155$, $0.12 \le \epsilon_{22} \le 1$, and $-0.03 \le \epsilon_{12} \le 0.03$. 

The mapping $\mathcal{F}: \mathbb{R}^6 \mapsto \mathbb{R}^6$ from the ambient space to the mapped space is performed by a single invertible layer consisting of three hidden \texttt{elu} layers with ten hidden units per layer. Neural network weights and biases are initialized by the uniform Kaiming approach.

\begin{figure}[h]
 \centering
{\includegraphics[width=0.5\textwidth]{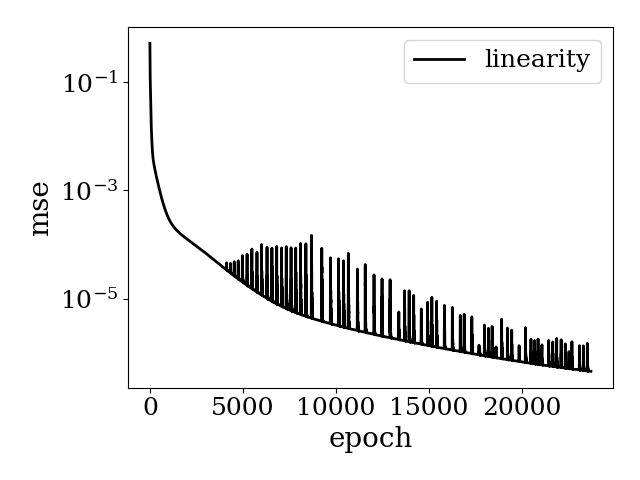}}
  \caption{Mean squared error (mse) of the linearity condition violation during the training epochs for the 2D nonlinear elasticity data. The training is performed in a full-batch manner with random shuffling at each epoch.
   \label{fig::pr4-loss-trn}}
\end{figure}

The neural network parameters are found by \texttt{ADAM} optimizer with an initial learning rate=0.002. We use \texttt{ReduceLROnPlateau} learning scheduler to adjust the learning rate every 50 iterations after the first 1000 iterations. The learning rate reduction factor is set to 0.91 with the minimum learning rate 1e-6. Before training, strain and stress data points are linearly normalized based on their maximum and minimum values to be positive and less than or equal to 1. The training performance is shown in \fig \ref{fig::pr4-loss-trn}.

The $\mathbb{C}$ parameter in the global optimization step is set to the Hessian of the strain energy at zero strain, and $\mathbb{S}$ tensor is set to its inverse. The same parameters are used in the local optimization for the original distance minimization method.

In \fig \ref{fig::pr4-fields}, we show the predicted displacement, strain, and stress contours by the introduced manifold method in this paper, the original distance minimization method, and the classical model-based method. Although the improvement in the displacement field predictions of the manifold approach over the nearest neighbor projection is marginal, the manifold method considerably enhances the strain and stress field predictions, 
both in terms of the magnitude of the errors and the symmetry of the solutions. 

The strain field obtained via the original distance minimization method (shown in \fig \ref{fig::pr4-fields} indicates that the discrete search does not yield a strain field that even qualitatively describes the overall characteristics of the problem, e.g., strain concentration around the hole imperfection. Notice that this is not unexpected since the amount of data used in this problem is very limited; 1000 data points to sample a 6-dimensional phase space is insufficient for the original distance minimization scheme.
Furthermore, it is also not difficult to see that the solutions obtained from the nearest neighbor projection may 
lose the symmetry due to the bias induced by the limited choice of data pointed. This limitation can be overcome or at least suppressed by the usage of embedded space for projection where each query point is projected onto a hyperplane instead of a point from the material database.

The proposed method is also conceptually different from the classical constitutive laws approach in which 
we merely \textit{implicitly} leverage the generalization power of deep neural networks in high-dimensional space (cf. \citet{balestriero2021learning, belkin2019reconciling})
to gain knowledge on the manifold and utilize this acquired knowledge to generate a more precise notion of distance. 
We did not explicitly introduce a specific form or equations to curve-fitting the data. As such, the resultant paradigm remains general and applicable for different mechanics problems.

\begin{table}[ht]
\centering
\begin{tabular}{m{2cm}m{4cm}m{4cm}m{4cm}}
{}&Displacement Norm&$\sqrt{3 J_2}$ of Strain Tensor&Von-misses Stress\\\\
Global Embedding Projection (this study)
&
\includegraphics[scale=.1]{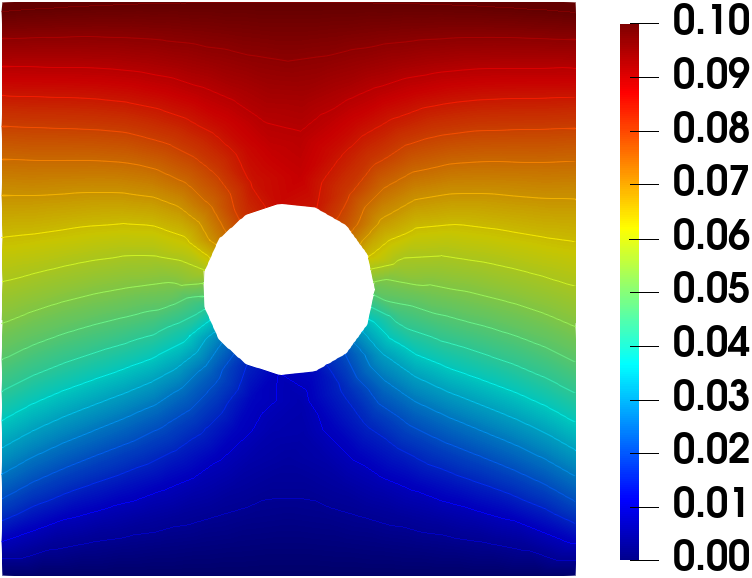}&
\includegraphics[scale=.1]{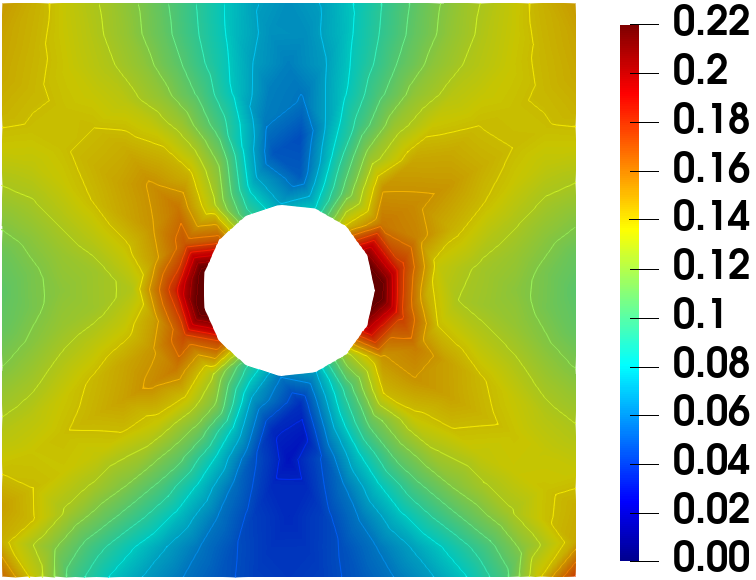}
&
\includegraphics[scale=.1]{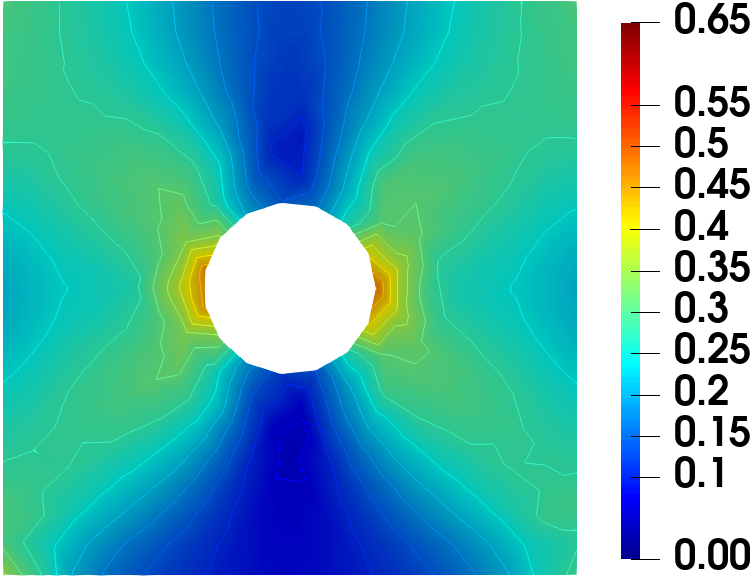}
\\
Nearest Neighbor Projection
&
\includegraphics[scale=.1]{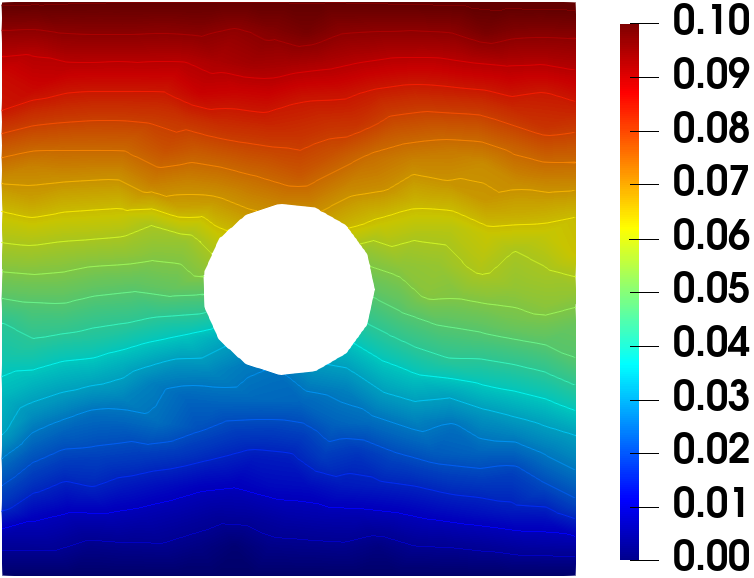}&
\includegraphics[scale=.1]{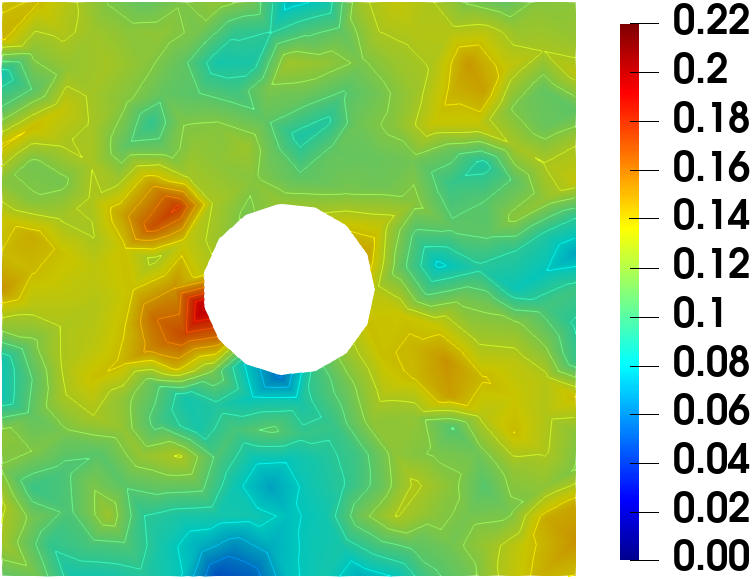}
&
\includegraphics[scale=.1]{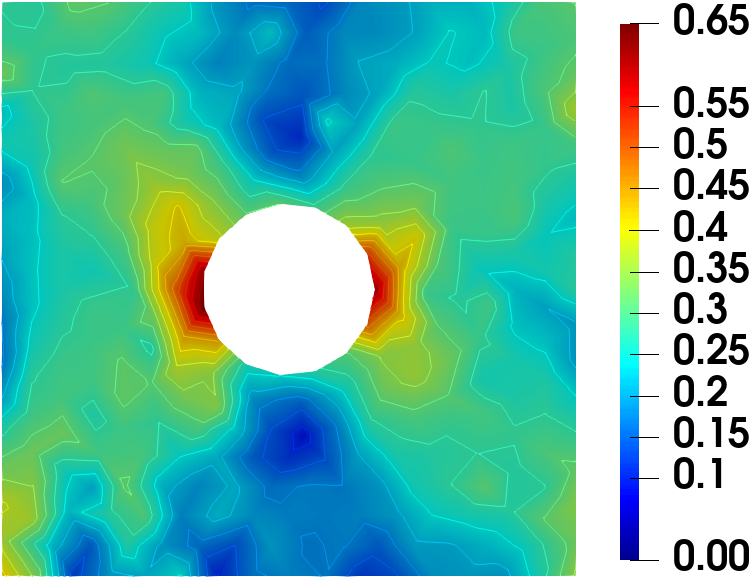}
\\
Classical FEM
&
\includegraphics[scale=.1]{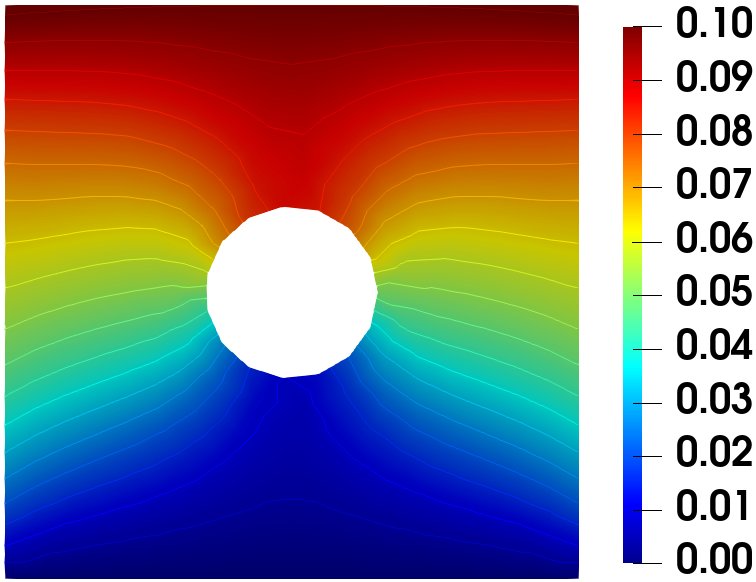}&
\includegraphics[scale=.1]{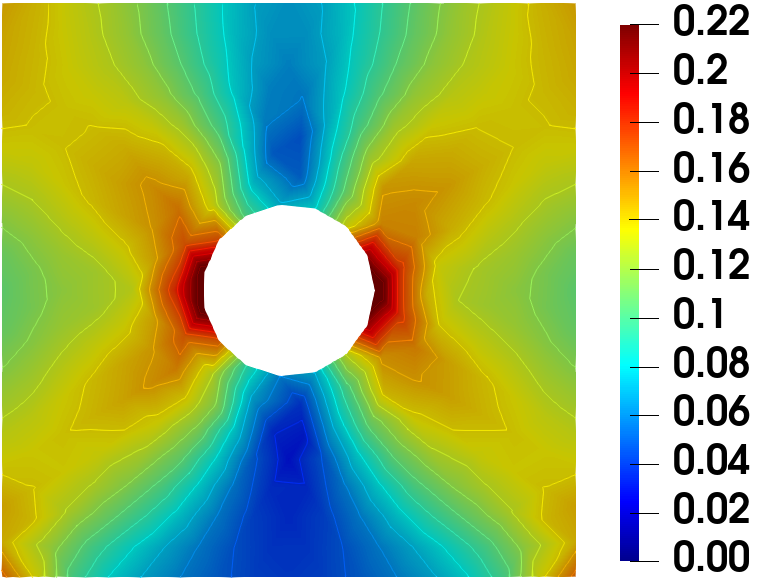}
&
\includegraphics[scale=.1]{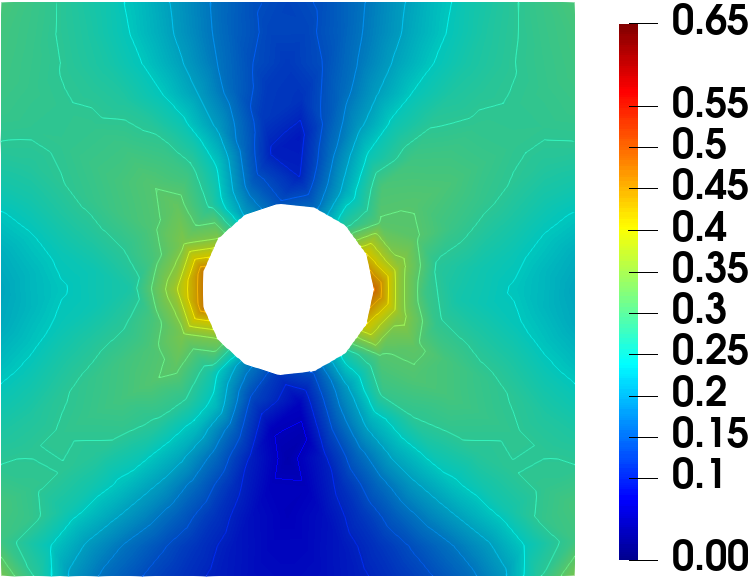}
\\
\end{tabular}
\captionof{figure}{A comparison between solution fields obtained by the two data-driven methods, i.e., global embedding (this study) and nearest neighbor projections, versus classical model-based nonlinear FEM (as ground truth). The simulations are performed under the plane strain condition.}
\label{fig::pr4-fields}
\end{table}

\begin{figure}[h]
 \centering
{\includegraphics[width=0.5\textwidth]{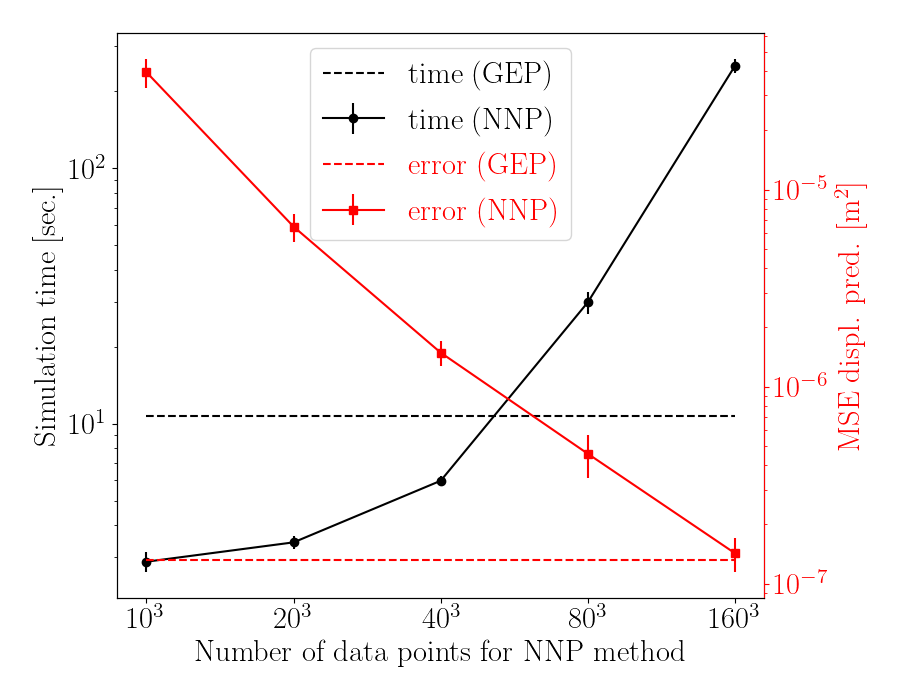}}
  \caption{simulation time (left axis) and displacement error (right axis) comparisons between the global embedding projection (GEP) and nearest neighbor projection (NNP) methods. The horizontal axis demonstrates the different amounts of data used in the NNP simulations. Vertical bars at each point show 1.5 standard deviations among 10 randomly initialized simulations for a fixed database. The markers indicate the average among these 10 simulations. For the GEP method, we use the database with $10^3$ data points, however, 10 randomly initialized simulations are also executed for this method. The average simulation time and displacement error for the GEP method are plotted by dashed black and red lines, respectively. The final solution by the GEP method is almost insensitive to the initialization (zero standard deviation in calculated errors). However, the mean and standard deviation values of simulation time for the GEP method are 10.7 and 0.3 seconds, respectively.
   \label{fig::pr4-time-err}}
\end{figure}

In \fig \ref{fig::pr4-time-err}, for the nearest neighbor projection method, the amount of data is gradually increased until we reach to almost similar error (for nodal displacements compared with classical model-based finite element) obtained by the global embedding projection method that utilizes a database consisting of only $10^3$ data points. Notice that (1) the classical approach needs 3 orders of magnitude more data than the proposed scheme to obtain a similar error in the displacement field prediction. (2) assuming the availability of this amount of data, the simulation time is  1 order of magnitude more than the proposed scheme here. (3) the proposed scheme is almost insensitive to the random initialization, which increases robustness.

\section{Conclusions} \label{sec::conclusion}
In this paper, we introduce a machine learning approach to discover the embedding vector space of constitutive manifolds and thereby improve the robustness, accuracy, and efficiency of the model-free paradigm originally proposed to completely bypass the need for constitutive laws. As demonstrated in the numerical examples, 
this manifold framework provides an effective remedy to enable the model-free paradigm to make reasonable predictions even with only a limited amount of data or in the cases where data are distributed unevenly. 
From the theoretical standpoint, the proposed method also introduces the notion of distance in a way more 
consistent with the geometry of the data. The numerical examples show that the knowledge of the geometry and topology of the manifold 
gained from the unsupervised learning that train the invertible neural network
 is capable of making robust and plausible predictions  especially when data supplied for the model-free simulations are limited. 
Nevertheless, it should be noted that, while treating the data points as elements of a Euclidean space equipped with an arbitrary norm 
could be ad-hoc
the machine learning procedure proposed here 
only construct an approximated embedding based on empirical evidences provided by the data points. 
Finding the true constitutive manifold as well as the proper embedding with and without assumptions on the smoothness are both non-trivial. 
There are also a few research directions that are important but have not yet been explored, such as the construction of isometric embedding to directly measure 
the geodesic distance between data points, the search of optimal hyperparameters for the offline training and the denoising 
strategy on manifold. 
Due to the bijective requirement, we did not study the proper way to denoise data in the constitutive manifold. 
These research directions will be pursue in the future but are considered out of the scope of this work.

\section{Acknowledgments}
The authors are supported by the National Science Foundation under grant contracts CMMI-1846875 and OAC-1940203, and
 the Dynamic Materials and Interactions Program from the Air Force Office of Scientific 
Research under grant contracts FA9550-21-1-0391 and FA9550-21-1-0027.
These supports are gratefully acknowledged. 
The views and conclusions contained in this document are those of the authors, 
and should not be interpreted as representing the official policies, either expressed or implied, 
of the sponsors, including the U.S. Government. 
The U.S. Government is authorized to reproduce and distribute reprints for 
Government purposes notwithstanding any copyright notation herein.

\bibliographystyle{plainnat}
\bibliography{main}

\end{document}